\documentclass[runningheads]{llncs}
\usepackage{graphicx}
\usepackage{amsmath,amssymb} %
\usepackage{color}
\usepackage[width=122mm,left=12mm,paperwidth=146mm,height=193mm,top=12mm,paperheight=217mm]{geometry}
\usepackage{tabularx}
\usepackage[shortlabels]{enumitem}
\usepackage{booktabs}
\usepackage{hyperref}

\def\bigO2{\mbox{${\cal O}$}}
\def\bigO{O}

\def\mL{\mathcal{L}}

\def\mY{\mathcal{Y}}
\def\1n{\mathbf{1}_n}
\def\0{\mathbf{0}}
\def\1{\mathbf{1}}

\def\a1{\mbox{\bf a}_1}
\def\a2{\mbox{\bf a}_2}
\def\a3{\mbox{\bf a}_3}
\def\a4{\mbox{\bf a}_4}

\newcommand{\norm}[1]{\left\lVert#1\right\rVert}

\newcolumntype{L}[1]{>{\raggedright\arraybackslash}p{#1}}
\newcolumntype{C}[1]{>{\centering\arraybackslash}p{#1}}
\newcolumntype{R}[1]{>{\raggedleft\arraybackslash}p{#1}}

\definecolor{forestgreen}{rgb}{0.23, 0.75, 0.02}

\newcommand*\samethanks[1][\value{footnote}]{\footnotemark[#1]}
\newcommand{\footref}[1]{\textsuperscript{\ref{#1}}}

\begin{document}
\pagestyle{headings}
\mainmatter

\title{A Style-Aware Content Loss for\\ Real-time HD Style Transfer}

\titlerunning{A Style-Aware Content Loss for Real-time HD Style Transfer}

\authorrunning{A. Sanakoyeu, D. Kotovenko, S. Lang, and B. Ommer}

\author{
Artsiom Sanakoyeu\thanks{Both authors contributed equally to this work.} \and Dmytro Kotovenko\samethanks \and Sabine Lang \and Bj{\"o}rn Ommer
}

\institute{Heidelberg Collaboratory for Image Processing,\\ IWR, Heidelberg University, Germany\\
	\email{firstname.lastname@iwr.uni-heidelberg.de}
}

\maketitle

\begin{abstract}
Recently, style transfer has received a lot of attention. While much of this research has aimed at speeding up processing, the approaches are still lacking from a principled, art historical standpoint: a style is more than just a single image or an artist, but previous work is limited to only a single instance of a style or shows no benefit from more images. Moreover, previous work has relied on a direct comparison of art in the domain of RGB images or on CNNs pre-trained on ImageNet, which requires millions of labeled object bounding boxes and can introduce an extra bias, since it has been assembled without artistic consideration. To circumvent these issues, we propose a style-aware content loss, which is trained jointly with a deep encoder-decoder network for real-time, high-resolution stylization of images and videos. We propose a quantitative measure for evaluating the quality of a stylized image and also have art historians rank patches from our approach against those from previous work. These and our qualitative results ranging from small image patches to megapixel stylistic images and videos show that our approach better captures the subtle nature in which a style affects content.\footnote{\vspace{-2pt}Project page: \url{https://compvis.github.io/adaptive-style-transfer}.\label{footnote:project_url}}
\end{abstract}

\section{Introduction}
\begin{figure}[!b]
    \centering
    \setlength{\tabcolsep}{-0.0pt}
    \begin{tabular}{c c c c c}
    \includegraphics[width=0.2\textwidth]{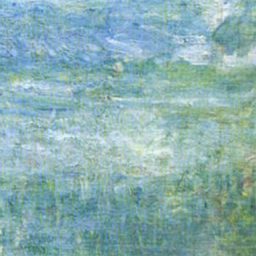} & 
    \includegraphics[width=0.2\textwidth]{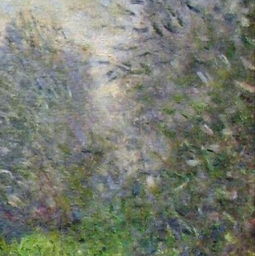} & 
    \includegraphics[width=0.2\textwidth]{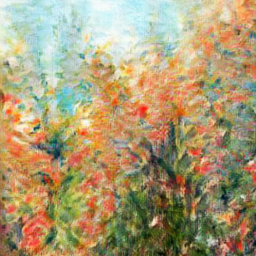} &
    \includegraphics[width=0.2\textwidth]{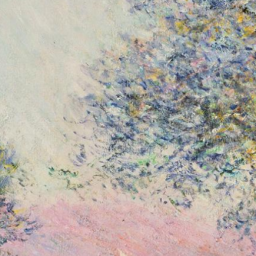} &
    \includegraphics[width=0.2\textwidth]{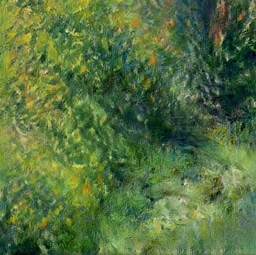}
    \end{tabular}
    \caption{Evaluating the fine details preserved by our approach. Can you guess which of the cut-outs are from Monet's artworks and which are generated? Solution is on p.~\pageref{trick:solution}.}
    \label{fig:trick}
\end{figure}

\begin{figure}[!t]
    \centering
    \def\arraystretch{0.1}
    \setlength{\tabcolsep}{-0.0pt}
    \begin{tabular}{cccccc}
     & \tiny{\cite{gatys2016}} & \tiny{\cite{gatys2016} } &
     \tiny{\cite{gatys2016} on collection} & \tiny{\cite{johnson} on collection} & \tiny{Ours on collection} \\
     
    Content & \includegraphics[width=0.16\textwidth,height=0.12\textwidth]{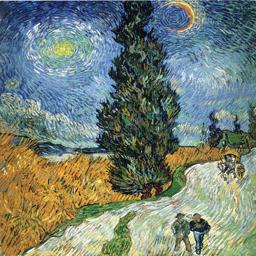} &
    \includegraphics[width=0.16\textwidth,height=0.12\textwidth]{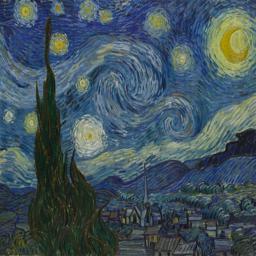} & 
    \includegraphics[width=0.16\textwidth,height=0.12\textwidth]{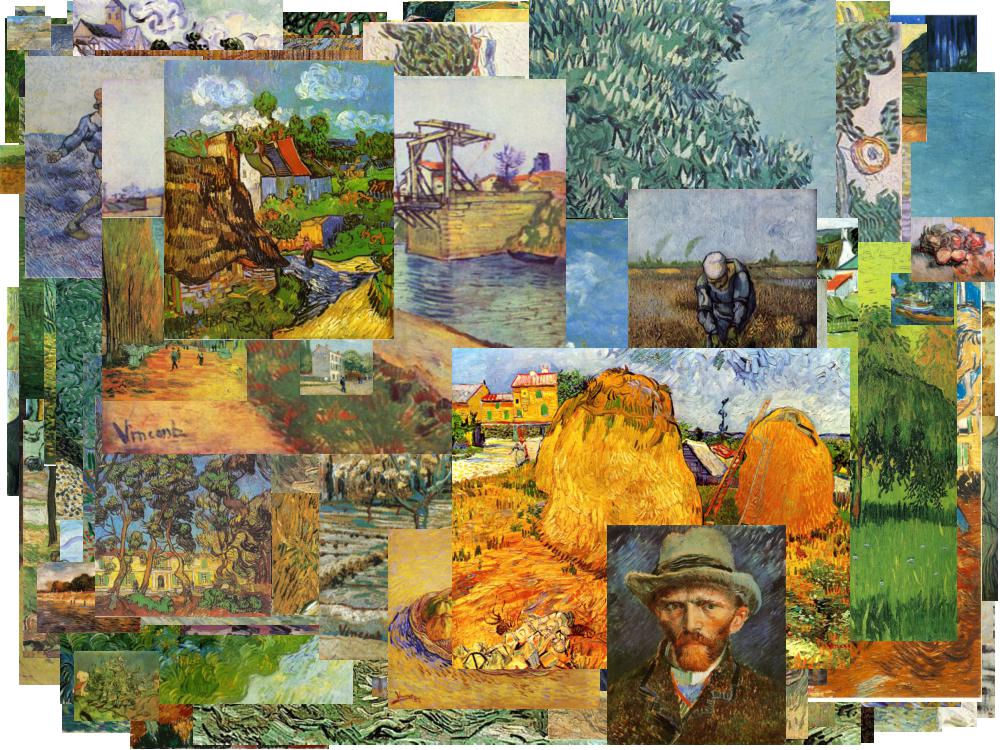} & 
    \includegraphics[width=0.16\textwidth,height=0.12\textwidth]{img/misc/collection_of_van-gogh_paintings.jpg} & 
    \includegraphics[width=0.16\textwidth,height=0.12\textwidth]{img/misc/collection_of_van-gogh_paintings.jpg} 
    \\

    \includegraphics[width=0.16\textwidth]{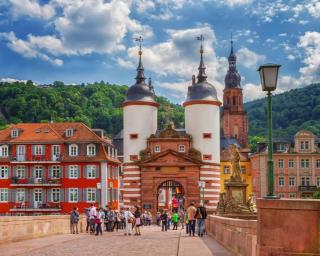} &
    \includegraphics[width=0.16\textwidth]{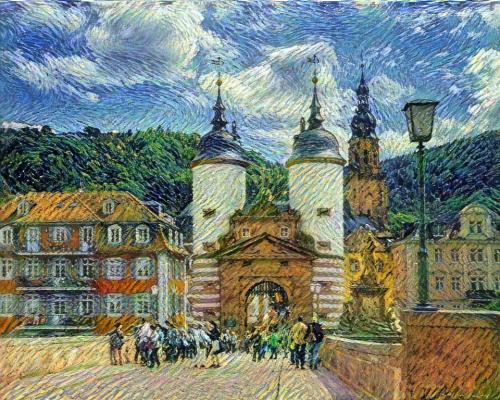} &
    \includegraphics[width=0.16\textwidth]{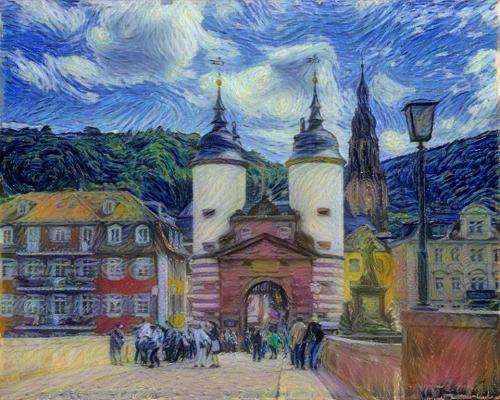} &
    \includegraphics[width=0.16\textwidth]{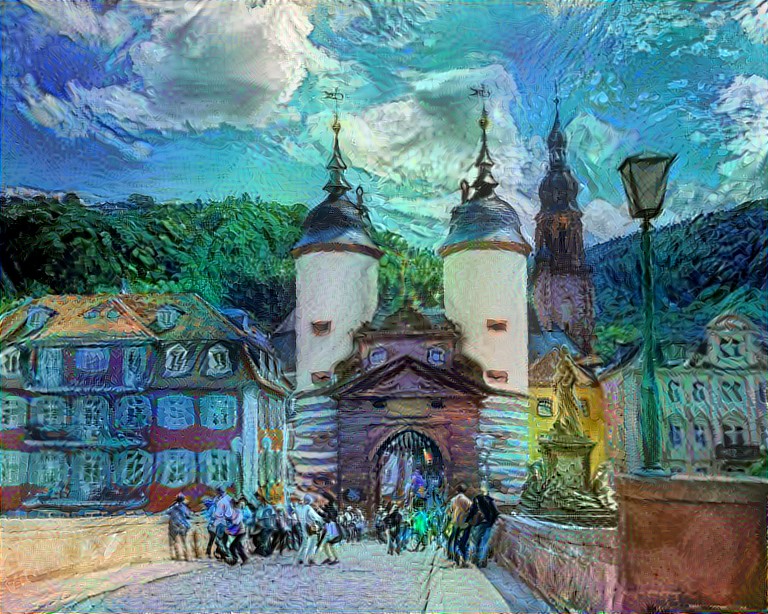}  &
    \includegraphics[width=0.16\textwidth]{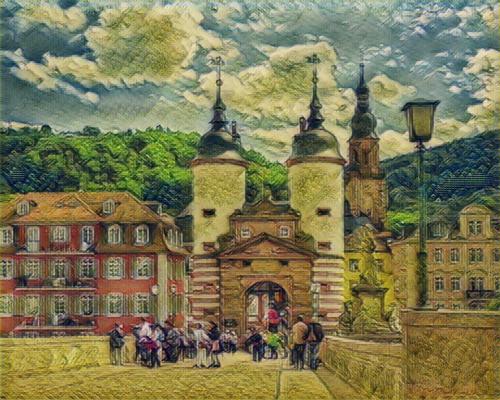} &
    \includegraphics[width=0.16\textwidth]{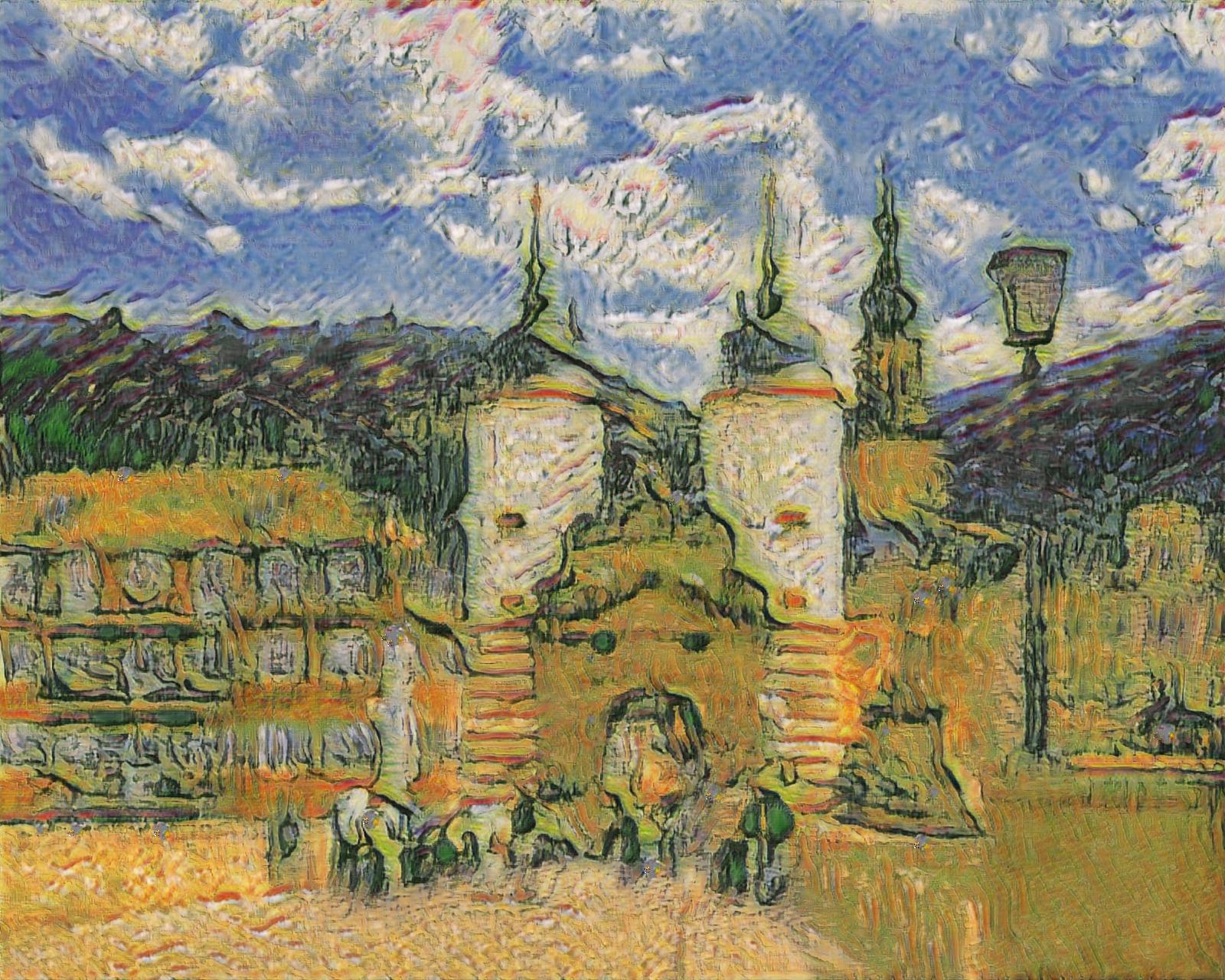} \\

    \includegraphics[width=0.16\textwidth]{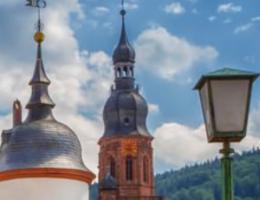} &
    \includegraphics[width=0.16\textwidth]{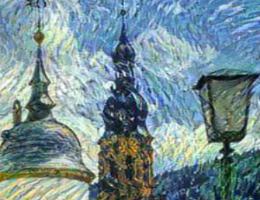} &
    \includegraphics[width=0.16\textwidth]{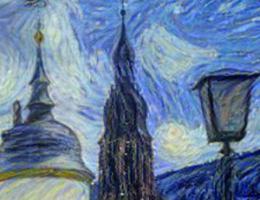} &
    \includegraphics[width=0.16\textwidth]{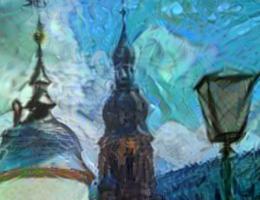} & 
    \includegraphics[width=0.16\textwidth]{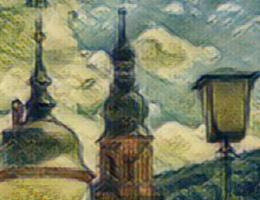} &
    \includegraphics[width=0.16\textwidth]{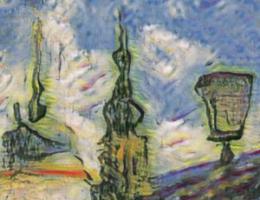} 
    \\

    & (a) & (b) & (c) & (d) &\textbf{(e)} \\

    \end{tabular}
    \caption{Style transfer using different approaches on $1$ and a collection of reference style images. 
    (a) \cite{gatys2016} using van Gogh's "Road with Cypress and Star" as reference style image; 
    (b) \cite{gatys2016} using van Gogh's "Starry night"; 
    (c) \cite{gatys2016} using the average Gram matrix computed across the collection of Vincent van Gogh's artworks; 
    (d) \cite{johnson} trained on the collection of van Gogh's artworks alternating target style images every SGD mini-batch;
    (e) our approach trained on the same collection of van Gogh's artworks.
    Stylizations \textit{(a)} and \textit{(b)} depend significantly on the particular style image, but using a collection of the style images \textit{(c)}, \textit{(d)}  does not produce visually plausible results, due to oversmoothing over the numerous Gram matrices. In contrast, our approach \textit{(e)} has learned how van Gogh is altering particular content in a specific manner (edges around objects also stylized, cf. bell tower)} 
    \label{fig:van_gogh_2ref}
\end{figure}

A picture may be worth a thousand words, but at least %
 it contains a lot of very diverse information. This not only comprises \emph{what} is portrayed, e.g., composition of a scene and individual objects, but also \emph{how} it is depicted, referring to the artistic style of a painting or filters applied to a photo. Especially when considering artistic images, it becomes evident that not only content but also style is a crucial part of the message an image communicates (just imagine van Gogh's Starry Night in the style of Pop Art). Here, we follow the common wording of our community and refer to 'content' as a synonym for 'subject matter' or 'sujet', preferably used in art history. A vision system then faces the challenge to decompose and separately represent the content and style of an image to enable a direct analysis based on each individually. The ultimate test for this ability is style transfer \cite{gatys2016} -- exchanging the style of an image while retaining its content.

\begin{figure}[t!]
    \centering
    \includegraphics[width=1.0\textwidth]{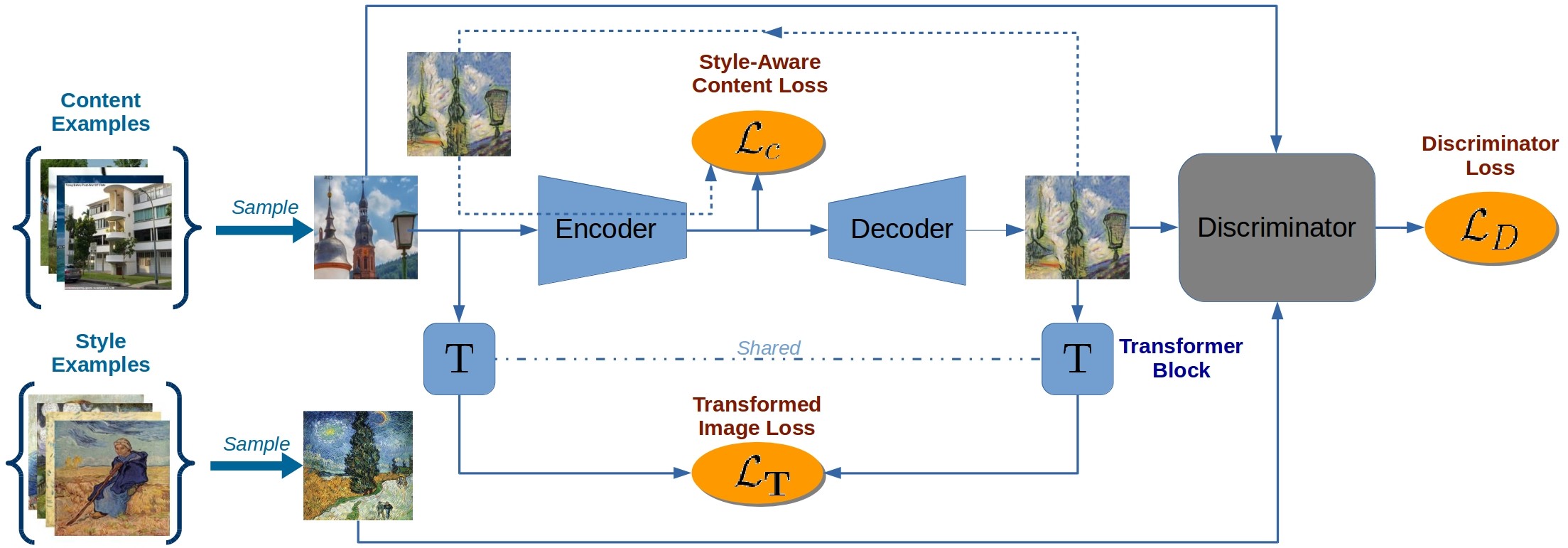}
    \caption{Encoder-decoder network for style transfer based on style-aware content loss.}
    \label{fig:pipeline}
\end{figure}

In contrast to the seminal work of Gatys et al. \cite{gatys2016}, who have relied on powerful but slow iterative optimization, there has recently been a focus on feed-forward generator networks \cite{johnson,multiscale_style,ulyanov2016texture,ulyanov2017improved,universal_style,dumoulin2016learned,adain}. The crucial representation in all these approaches has been based on a VGG16 or VGG19 network \cite{vgg}, pre-trained on ImageNet \cite{imagenet}. However, a recent trend in deep learning has been to avoid supervised pre-training on a million images with tediously labeled object bounding boxes \cite{wang2015unsupervised}. In the setting of style transfer this has the particular benefit of avoiding from the outset any bias introduced by ImageNet, which has been assembled without artistic consideration. 
Rather than utilizing a separate pre-trained VGG network to measure and optimize the quality of the stylistic output \cite{gatys2016,johnson,multiscale_style,ulyanov2016texture,ulyanov2017improved,universal_style,dumoulin2016learned}, we employ an encoder-decoder architecture with adversarial discriminator, Fig. \ref{fig:pipeline}, to stylize the input content image and also use the encoder to measure the reconstruction loss. In essence the stylized output image is again run through the encoder and compared with the encoded input content image. Thus, we learn a style-specific content loss from scratch, which adapts to the specific way in which a particular style retains content and is more adaptive than a comparison in the domain of RGB images \cite{cyclegan}.

Most importantly, however, previous work has only been based on a \emph{single} style image. This stands in stark contrast to art history 
which understands "style as an expression of a collective spirit" 
resulting in a "distinctive manner which permits the grouping of works into related categories"~\cite{fernie}.
As a result, art history developed a scheme, which allows to identify groups of artworks based on shared qualities. Artistic style consists of a diverse range of elements, such as form, color, brushstroke, or use of light. Therefore, it is insufficient to only use a single artwork, because it might not represent the full scope of an artistic style. Today, freely available art datasets such as Wikiart \cite{karayev2013recognizing} easily contain more than 100K images, thus providing numerous examples for various styles. Previous work \cite{gatys2016,johnson,multiscale_style,ulyanov2016texture,ulyanov2017improved,universal_style,dumoulin2016learned} has represented style based on the Gram matrix, which captures highly image-specific style statistics, cf. Fig.~\ref{fig:van_gogh_2ref}.
To combine several style images in \cite{gatys2016,johnson,multiscale_style,ulyanov2016texture,ulyanov2017improved,universal_style,dumoulin2016learned} one needs to aggregate their Gram matrices. We have evaluated several aggregation strategies and averaging worked the best, Fig.~\ref{fig:van_gogh_2ref}(c). But, obviously, neither art history, nor statistics suggests aggregating Gram matrices. Additionally, we investigated alternating the target style images in every mini-batch while training \cite{johnson}, Fig.~\ref{fig:van_gogh_2ref}(d). However, all these methods cannot make proper use of several style images, because combining the Gram matrices of several images forfeits the details of style, cf. the analysis in Fig.~\ref{fig:van_gogh_2ref}.  
In contrast, our proposed approach allows to combine an arbitrary number of instances of a style during training.

We conduct extensive evaluations of the proposed style transfer approach; we quantitatively and qualitatively compare it against numerous baselines. Being able to generate high quality artistic works in
high-resolution, our approach produces visually more detailed stylizations than the current state of the art style transfer approaches and yet shows real-time inference speed. The results are quantitatively validated by experts from art history and by adopting in this paper a \emph{deception rate} metric  based on a deep neural network for artist classification.

\subsection{Related Work}
In recent years, a lot of research efforts have been devoted to texture synthesis and style transfer problems. 
Earlier methods \cite{hertzmann2001image} are usually non-parametric and are build upon low-level image features. 
Inspired by Image Analogies~\cite{hertzmann2001image},  approaches \cite{split_n_match,liao2017_deep_image_analogy,style_portraits,day_night_hallucination} are based on finding dense correspondence between content and style image and often require image pairs to depict similar content. Therefore, these methods do not scale to the setting of arbitrary content images.

In contrast, Gatys et al.~\cite{gatys2016,gatys2015texture} proposed a more flexible iterative optimization approach based on a pre-trained VGG19 network \cite{vgg}. This method produces high quality results and works on arbitrary inputs, but is costly, since each optimization step requires a forward and backward pass through the VGG19 network. 
Subsequent methods \cite{johnson,ulyanov2016texture,markov_net} aimed to accelerate the optimization procedure \cite{gatys2016} by approximating it with feed-forward convolutional neural networks.
This way, only one forward pass through the network is required to generate a stylized image. 
Beyond that, a number of methods have been proposed to address different aspects of style transfer, including quality \cite{gatys2017controlling,hist_loss_2017,pb,multiscale_style,jing2018stroke}, diversity \cite{diversified_2017,ulyanov2017improved}, photorealism \cite{deep_photo_style}, combining several styles in a single model \cite{zm_net,dumoulin2016learned,stylebank} and generalizing to previously unseen styles \cite{adain,universal_style,exploring_bmvc,shen2017meta}. However, all these methods rely on the fixed style representation which is captured by the features of a VGG~\cite{vgg} network pre-trained on ImageNet. %
Therefore they require a supervised pre-training on millions of labeled object bounding boxes and have a bias introduced by ImageNet, because it has been assembled without artistic consideration. 
Moreover, the image quality achieved by the costly optimization in \cite{gatys2016} still remains an upper bound for the performance of recent methods. 
Other works like \cite{Wilber_2017_ICCV,Collomosse_2017_ICCV,deepart,esser2018variational,cliquecnn} learn how  to discriminate different techniques, styles and contents in the latent space.
Zhu et al.~\cite{cyclegan} learn a bidirectional mapping between 
a domain of content images and paintings using generative adversarial networks. Employing cycle consistency loss, they directly measure the distance between a backprojection of the stylized output and the content image in the RGB pixel space.
Measuring distances in the RGB image domain is not just generally prone to be coarse, but, especially for abstract styles, a pixel-wise comparison of backwards mapped stylized images is not suited.
Then, either content is preserved and the stylized image is
not sufficiently abstract, e.g., not altering object boundaries, or the stylized image has a suitable degree of abstractness and so a pixel-based comparison with the  content  image  must  fail.  Moreover,  the  more  abstract  the  style is,  the  more
potential  backprojections  into  the  content  domain  exist,  because  this mapping is underdetermined (think of the many possible content images for a single cubistic painting). 
In contrast, we spare the ill-posed backward mapping of styles and compare stylized and content images in the latent space which is trained jointly with the style transfer network.
Since both content and stylized images are run through our encoder, the latent space is trained to only pay attention to the commonalities, i.e., the content present in both.
Another consequence of the cycle consistency loss is that it requires content and style images used for training to represent similar scenes \cite{cyclegan}, and thus training data preparation for \cite{cyclegan} involves tedious manual filtering of samples, while our approach can be trained on arbitrary unpaired content and style images.

\section{Approach}
\label{sec:approach}

\begin{figure}[t!]
    \centering
    \def\arraystretch{0.0}
    \setlength{\tabcolsep}{-0.0pt}
    \begin{tabular}{ccccc}
    Content & (a) Pollock & (b) El-Greco & (c) Gauguin & (d) C\'ezanne\\
    \noalign{\smallskip} & \\
    
    \includegraphics[width=0.2\textwidth,height=0.15\textwidth]{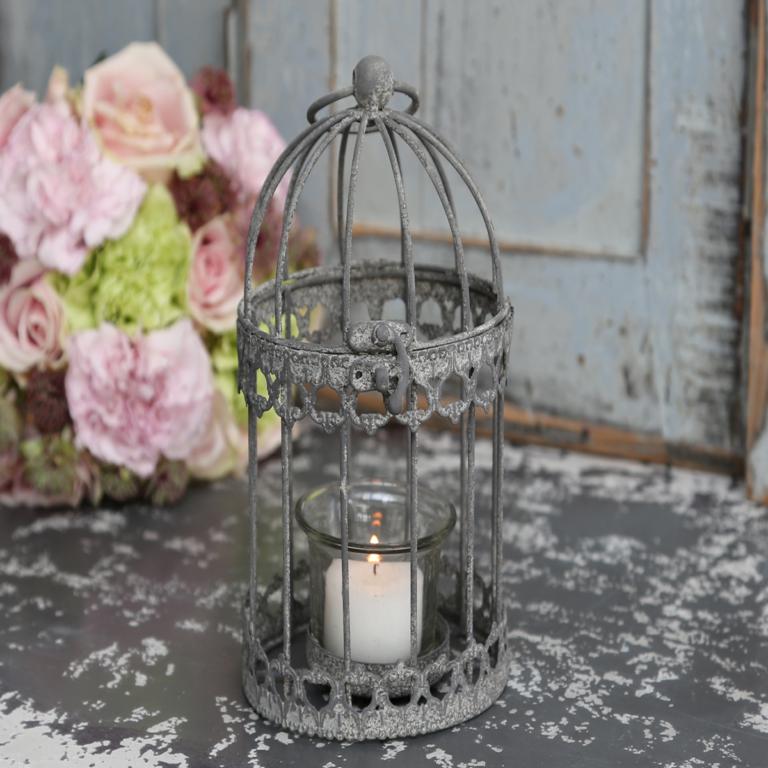} &
    \includegraphics[width=0.2\textwidth,height=0.15\textwidth]{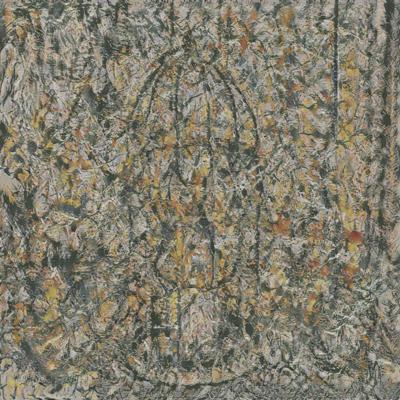} &
    \includegraphics[width=0.2\textwidth,height=0.15\textwidth]{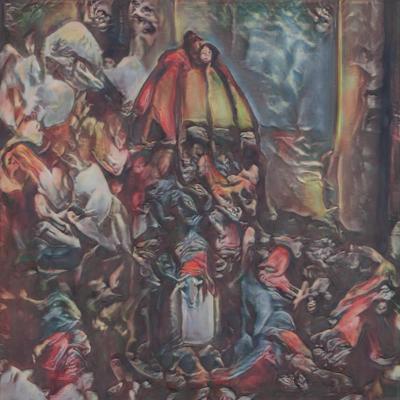} &
    \includegraphics[width=0.2\textwidth,height=0.15\textwidth]{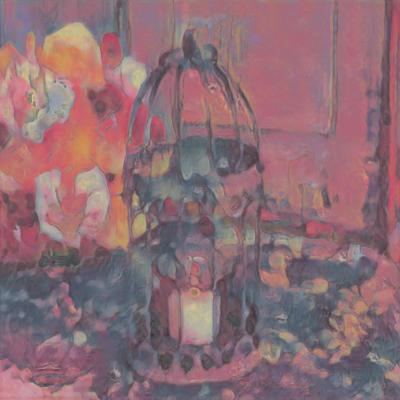} &
    \includegraphics[width=0.2\textwidth,height=0.15\textwidth]{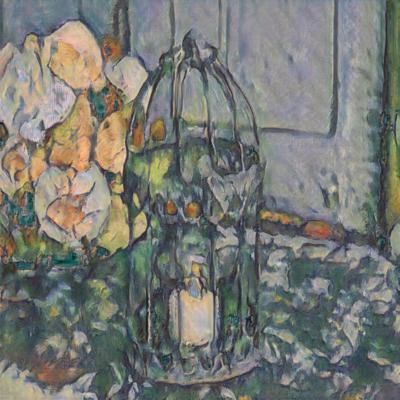}\\ 
    
    \includegraphics[width=0.2\textwidth,height=0.15\textwidth]{img/misc/reconstruction/60819-00_original.jpg} &
    \includegraphics[width=0.2\textwidth,height=0.15\textwidth]{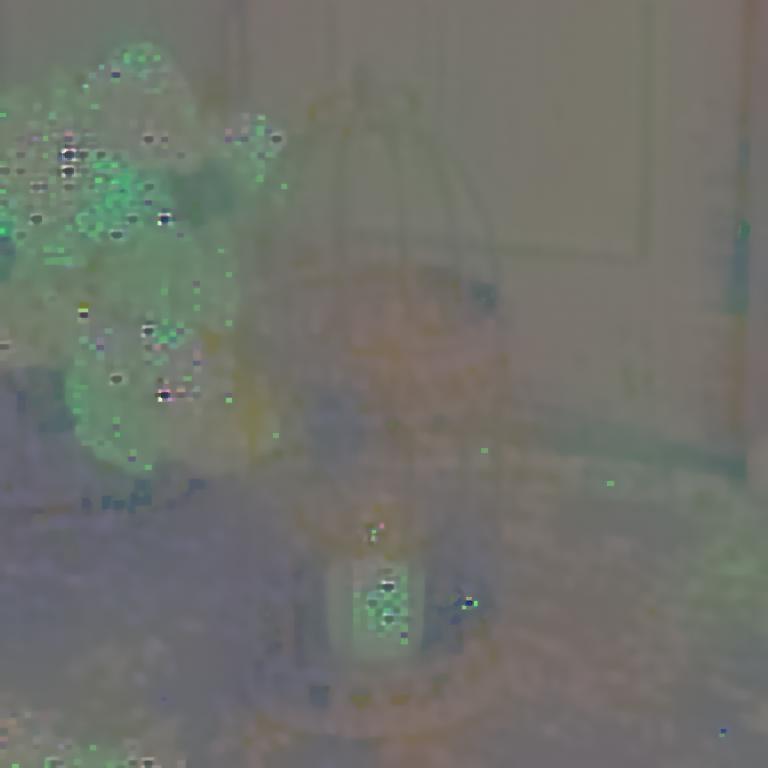} &
    \includegraphics[width=0.2\textwidth,height=0.15\textwidth]{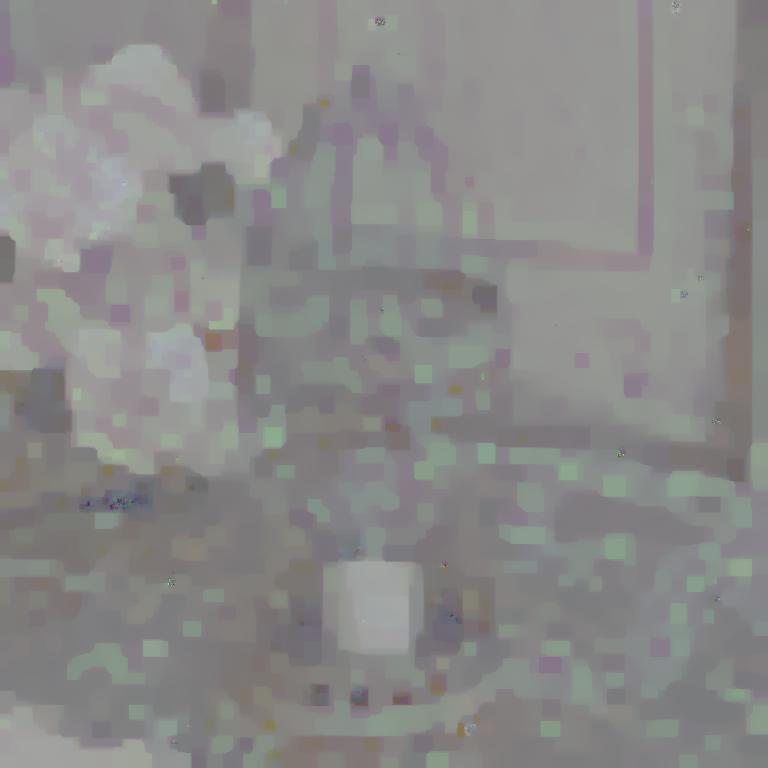} &
    \includegraphics[width=0.2\textwidth,height=0.15\textwidth]{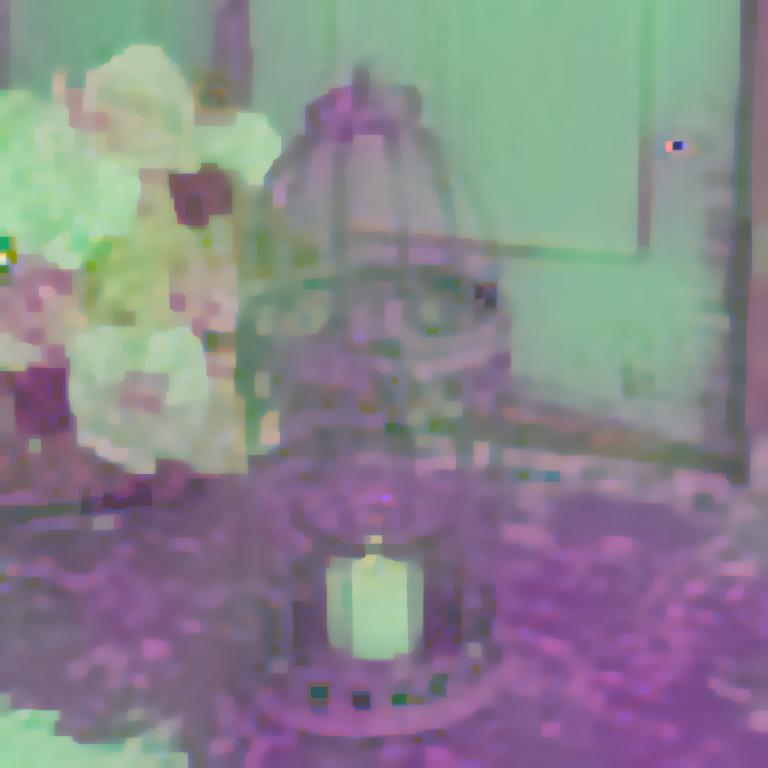} &
    \includegraphics[width=0.2\textwidth,height=0.15\textwidth]{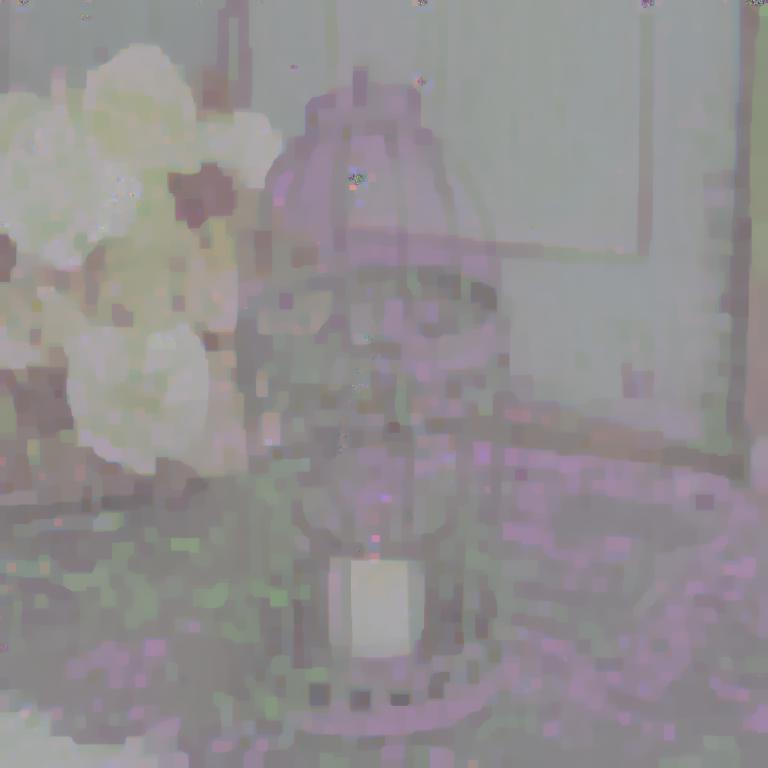} 
    
    \end{tabular}

    \caption{$1$st row - results of style transfer for different styles. $2$nd row - sketchy content visualization reconstructed from the latent space $E(x)$ using method of \cite{vedaldi}. (a) The encoder for Pollock does not preserve much content due to the abstract style; (b) only rough structure of the content is preserved (coarse patches) because of the distinct style of El Greco; (c) latent space highlights surfaces of the same color and that fine object details are ignored, since Gauguin was less interested in details, often painted plain surfaces and used vivid colors; (d) encodes the thick, wide brushstrokes C\'ezanne used, but preserves a larger palette of colors. }
    \label{fig:reconstruction}
\end{figure}

To enable a fast style transfer that instantly transfers a content image or even frames of a video according to a particular style, we need a feed-forward architecture \cite{johnson} rather than the slow optimization-based approach of \cite{gatys2016}. To this end, we adopt an encoder-decoder architecture that utilizes an encoder network $E$ to map an input content image $x$ onto a latent representation $z=E(x)$. A generative decoder $G$ then plays the role of a painter and generates the stylized output image $y=G(z)$ from the sketchy content representation $z$. Stylization then only requires a single forward pass, thus working in real-time.

\subsection{Training with a Style-Aware Content Loss}

Previous approaches have been limited in that training worked only with a single style image \cite{gatys2016,johnson,adain,universal_style,multiscale_style,dumoulin2016learned,ulyanov2016texture} or that style images used for training had to be similar in content to the content images \cite{cyclegan}. In contrast, given a single style image $y_0$ we include a set $Y$ of related style images $y_j \in Y$, which are automatically selected (see Sec.~\ref{sec:grouping}) from a large art dataset (Wikiart). We do \emph{not} require the $y_j$ to depict similar content as the set $X$ of arbitrary content images $x_i \in X$, which we simply take from Places365 \cite{places365}. Compared to \cite{cyclegan}, we thus can utilize standard datasets for content and style and need no tedious manual selection of the $x_i$ and $y_j$ as described in Sect. 5.1 and 7.1 of \cite{cyclegan}.

To train $E$ and $G$ we employ a standard adversarial discriminator $D$ \cite{gans} to distinguish the stylized output $G(E(x_i))$ from real examples $y_j\in Y$,

\begin{equation}
\label{eq:gan_loss}
\mathcal{L}_D(E, G, D) =   \mathop{\mathbb{E}}_{y \sim p_{Y}(y)} \left[ \log{D(y)} \right] + \mathop{\mathbb{E}}_{x \sim p_{X}(x)} \left[ \log{\left(1 - D(G(E(x)))\right)} \right]
\end{equation}

However, the crucial challenge is to decide which details to retain from the content image, something which is not captured by Eq.~\ref{eq:gan_loss}. Contrary to previous work, we want to directly enforce $E$ to strip the latent space of all image details that the target style disregards. Therefore, the details that need to be retained or ignored in $z$ depend on the style. For instance, Cubism would disregard texture, whereas Pointillism would retain low-frequency textures. Therefore, a pre-trained network or fixed similarity measure \cite{gatys2016} for measuring the similarity in content between $x_i$ and $y_i$ is violating the art historical premise that the manner, in which content is preserved, depends on the style. Similar issues arise when measuring the distance after projecting the stylized image $G(E(x_i))$ back into the domain $X$ of original images with a second pair of encoder and decoder $G_2(E_2(G(E(x_i))))$. The resulting loss proposed in \cite{cyclegan},
\begin{equation}
    \mL_{cycleGAN} = \mathop{\mathbb{E}}_{x \sim p_{X}(x)} \left[ \norm{ x - G_2(E_2(G(E(x))) )}_1 \right],
\end{equation}
fails where styles become abstract, since the backward projection of abstract art to the original image is highly underdetermined.

Therefore, we propose a style-aware content loss that is being optimized, while the network learns to stylize images. %
Since encoder training is coupled with training of the decoder, which produces artistic images of the specific style,
the latent vector $z$ produced for the input image $x$ can be viewed as its style-dependent sketchy content representation. This latent space representation is changing during training and hence \emph{adapts} to the style.
Thus, when measuring the similarity in content between input image $x_i$ and the stylized image $y_i = G(E(x_i))$ in the latent space, we focus only on those details which are relevant for the style.
Let the latent space have $d$ dimensions, then we define a style-aware content loss as normalized squared Euclidean distance between $E(x_i)$ and $E(y_i)$:
\begin{equation}
    \mL_{c}(E, G) = \mathop{\mathbb{E}}_{x \sim p_{X}(x)} \left[ \frac{1}{d} \norm{ E(x) - E(G(E(x))) }_2^2 \right]
\end{equation}
To show the additional intuition behind the style-aware content loss we used the method \cite{vedaldi} to reconstruct the content image from latent representations trained on different styles and illustrated it in Fig.~\ref{fig:reconstruction}. It can be seen that latent space encodes a sketchy, style-specific visual content, 
which is implicitly used by the loss function.
For example, Pollock is famous for his abstract paintings, so reconstruction (a) shows that the latent space ignores most of the object structure; Gauguin was less interested in details, painted a lot of plain surfaces and used vivid colors which is reflected in the reconstruction (c), where latent space highlights surfaces of the same color and fine object details are ignored.

Since we train our model for altering the artistic style without supervision and from scratch, we now introduce extra signal to initialize training and boost the learning of the primary latent space. 
The simplest thing to do is to use an autoencoder loss 
which computes the difference between $x_i$ and $y_i$ in the RGB space.
However, this loss would impose a high penalty for any changes in image structure between input $x_i$ and output $y_i$,
because it relies only on low-level pixel information. 
But we aim to learn image stylization and want the encoder to discard certain details in the content depending on style.
Hence the autoencoder loss will contradict with the purpose of the style-aware loss,
where the style determines which details to retain and which to disregard.
Therefore, we propose to measure the difference after applying a weak image transformation on $x_i$ and $y_i$, which is learned while learning E and G. 
We inject in our model a transformer block $\mathbf{T}$ 
which is essentially a one-layer fully convolutional neural network taking an image as input and producing a transformed image of the same size. We apply $T$ to images $x_i$ and $y_i=G(E(x_i))$ before measuring the difference. We refer to this as \emph{transformed image loss} and define it as

\begin{equation}
\mL_{\mathbf{T}}(E, G) = \mathop{\mathbb{E}}_{x \sim p_{X}(x)} \left[ \frac{1}{C H W} || \mathbf{T}(x) - \mathbf{T}(G(E(x)) ||_2^2 \right],
\end{equation}
where $C \times H \times W$ is the size of image $x$ and for training $T$ is initialized with uniform weights.

Fig.~\ref{fig:pipeline} illustrates the full pipeline of our approach. To summarize, the full objective of our model is:
\begin{equation}
   \label{eq:total_loss}
    \mL(E, G, D) = \mL_c(E, G) + \mL_{t}(E, G) + \lambda \mL_D(E, G, D) ,
\end{equation}
where $\lambda$ controls the relative importance of adversarial loss. 
We solve the following optimization problem:
\begin{equation}
    E, G = \arg\;\min\limits_{E,G}\;\max\limits_{D} \mL(E, G, D).
\end{equation}

\subsection{Style Image Grouping}
\label{sec:grouping}

In this section we explain an automatic approach for gathering a set of related style images.
Given a single style image $y_0$ we strive to find a set $Y$ of related style images $y_j \in Y$.
Contrary to \cite{cyclegan} we avoid tedious manual selection of style images and follow a fully automatic approach. 
To this end, we train a VGG$16$~\cite{vgg} network $C$ from scratch on the Wikiart~\cite{karayev2013recognizing} dataset to predict an artist given the artwork. The network is trained on the $624$ largest (by number of works) artists from the Wikiart dataset.
Note that our ultimate goal is stylization and numerous artists can share the same style, e.g., Impressionism, as well as a single artist can exhibit different styles, such as the different stylistic periods of Picasso. However, we do \emph{not} use any \emph{style} labels. Artist classification in this case is the surrogate task for learning meaningful features in the artworks' domain, which allows to retrieve similar artworks to image $y_0$.

Let $\phi(y)$ be the activations of the fc$6$ layer of the VGG$16$ network $C$ for input image $y$. To get a set of related style images to $y_0$ from the Wikiart dataset $\mY$  we retrieve all nearest neighbors of $y_0$  based on the cosine distance $\delta$ of the activations $\phi(\cdot)$, i.e. 
\begin{equation}
    Y = \{ y  \; |\; y \in \mY, \delta(\phi(y), \phi(y_0)) < t \},
\end{equation}
where $\delta(a, b) = 1 + \frac{\phi(a)\phi(b)}{||a||_2 ||b||_2}$ and $t$ is the $10\%$ quantile of all pairwise distances in the dataset $\mY$.

\section{Experiments}
To compare our style transfer approach with the state-of-the-art, we first perform extensive qualitative analysis, then we provide quantitative results based on the \emph{deception score} and evaluations of experts from art history. Afterwards in Sect.~\ref{sec:ablation} we ablate single components of our model and show their importance.
\\[1mm]\textbf{Implementation details:}
The basis for our style transfer model is an encoder-decoder architecture, cf.~\cite{johnson}.
The encoder network contains $5$ \texttt{conv} layers: \\
$1 \times$\texttt{conv-stride-1} and $4\times$\texttt{conv-stride-2}. The decoder network has $9$ residual blocks \cite{he2016deep}, $4$ upsampling blocks and $1 \times$\texttt{conv-stride-1}. For upsampling blocks we used a sequence of nearest-neighbor upscaling and \texttt{conv-stride-$1$} instead of fractionally strided convolutions \cite{fully_conv}, which tend to produce heavier artifacts \cite{odena2016checkerboard}. Discriminator is a fully convolutional network with $7\times$\texttt{conv-stride-2} layers.
For a detailed network architecture description we refer to the supplementary material. 
We set $\lambda=0.001$ in Eq.~\ref{eq:total_loss}.
During the training process we sample $768 \times 768$ content image patches from the training set of Places365 \cite{places365} and $768 \times 768$ style image patches from the Wikiart \cite{karayev2013recognizing} dataset. We train for $300000$ iterations with batch size $1$, learning rate $0.0002$ and Adam \cite{adam} optimizer.  The learning rate is reduced by a factor of $10$ after $200000$ iterations.
\\[1mm]\textbf{Baselines:}
Since we aim to generate high-resolution stylizations, for comparison we run style transfer on our method and all baselines for input images of size $768\times 768$, unless otherwise specified. We did not not exceed this resolution when comparing, because some other methods were reaching  the GPU memory limit.  
We optimize Gatys et al.~\cite{gatys2016} for $500$ iterations using L-BFGS.
For Johnson et al.~\cite{johnson} we used the implementation of \cite{engstrom2016faststyletransfer} and trained a separate network for every reference style image on the same content images from Places365 \cite{places365} as our method.
For Huang et al. \cite{adain}, Chen et al. \cite{pb} and Li et al. \cite{universal_style} implementations and pre-trained models provided by the authors were used.
Zhu et al. \cite{cyclegan} was trained on exactly the same content and style images as our approach using the source code provided by the authors. Methods \cite{gatys2016,johnson,adain,pb,universal_style} utilized only one example per style, as they cannot benefit from more (cf. the analysis in Fig.~\ref{fig:van_gogh_2ref}).

\subsection{Qualitative Results}
\begin{figure}[!t]
    \centering
    \def\arraystretch{1}
    \setlength{\tabcolsep}{-0.0pt}
    
    \begin{tabular}{*{2}{C{0.11\textwidth}}*{7}{C{0.11\textwidth}}}
     Style & Content & (a) Ours & (b) \cite{gatys2016} & (c) \cite{cyclegan} & (d) \cite{pb} & (e) \cite{adain} & (f) \cite{universal_style} & (g) \cite{johnson}
    \end{tabular}
    
    \begin{tabular}{cc||cl}

    \includegraphics[width=0.11\textwidth,height=0.09\textwidth]{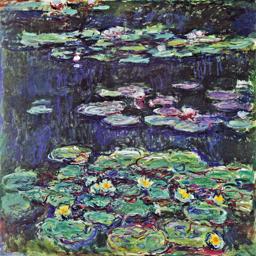} &
    \includegraphics[width=0.11\textwidth,height=0.09\textwidth]{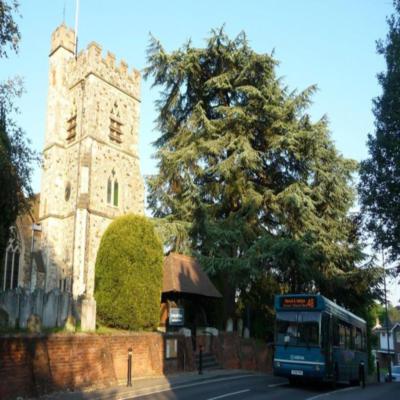} &
    \includegraphics[width=0.11\textwidth,height=0.09\textwidth]{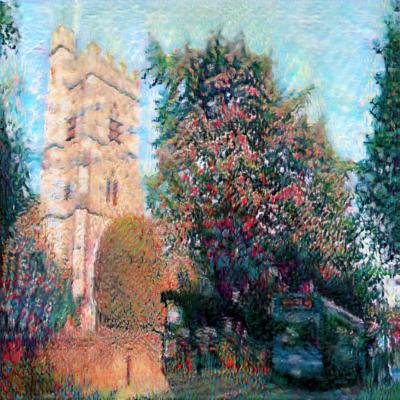} &
    \includegraphics[width=0.66\textwidth,height=0.09\textwidth]{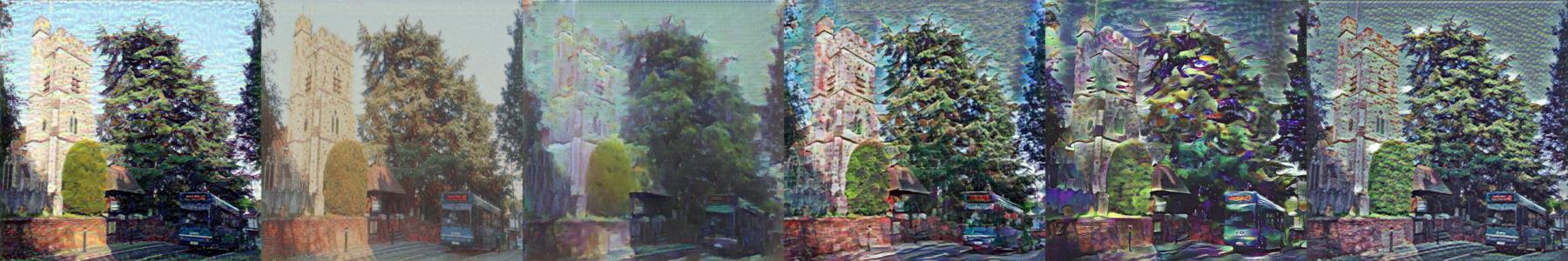}
    \\

    \includegraphics[width=0.11\textwidth,height=0.09\textwidth]{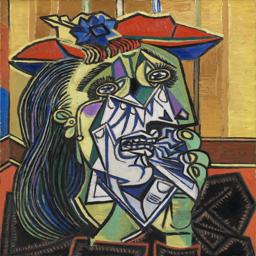} &
    \includegraphics[width=0.11\textwidth,height=0.09\textwidth]{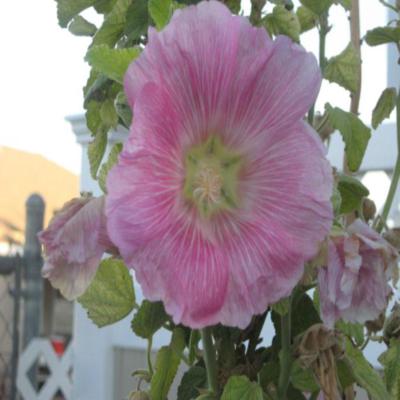} &
    \includegraphics[width=0.11\textwidth,height=0.09\textwidth]{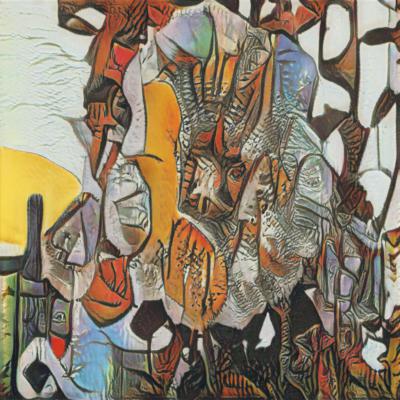} &
    \includegraphics[width=0.66\textwidth,height=0.09\textwidth]{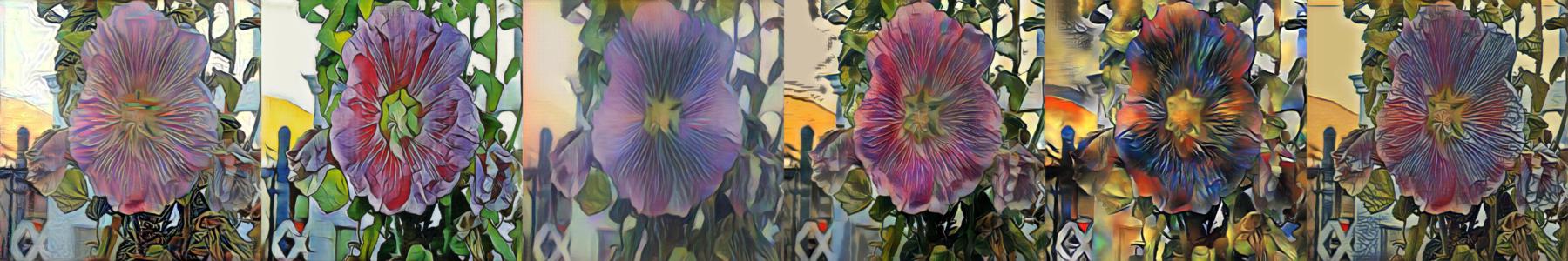}
    \\

    \includegraphics[width=0.11\textwidth,height=0.09\textwidth]{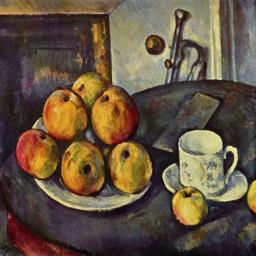} &
    \includegraphics[width=0.11\textwidth,height=0.09\textwidth]{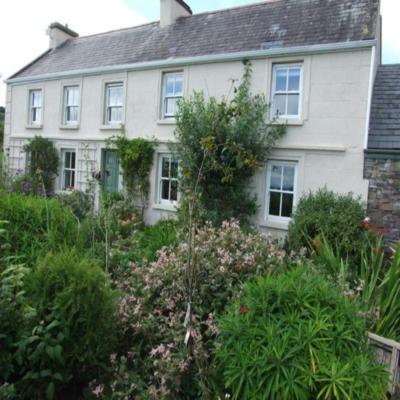} &
    \includegraphics[width=0.11\textwidth,height=0.09\textwidth]{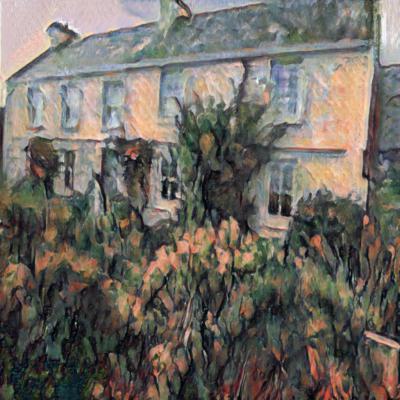} &
    \includegraphics[width=0.66\textwidth,height=0.09\textwidth]{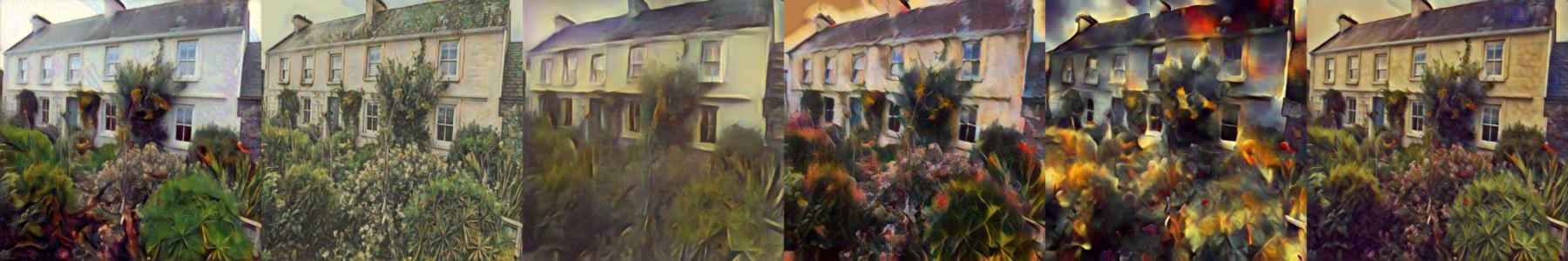}
    \\

    \includegraphics[width=0.11\textwidth,height=0.09\textwidth]{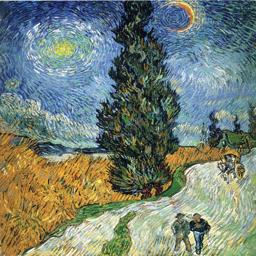} &
    \includegraphics[width=0.11\textwidth,height=0.09\textwidth]{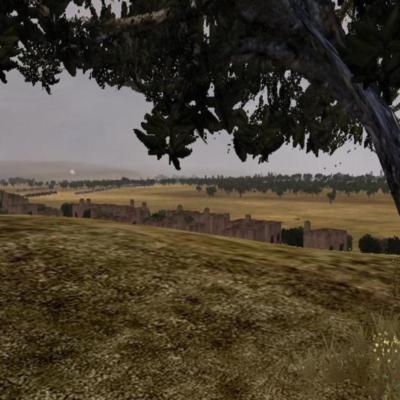} &
    \includegraphics[width=0.11\textwidth,height=0.09\textwidth]{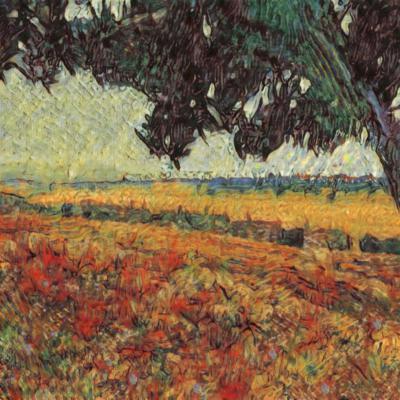} &
    \includegraphics[width=0.66\textwidth,height=0.09\textwidth]{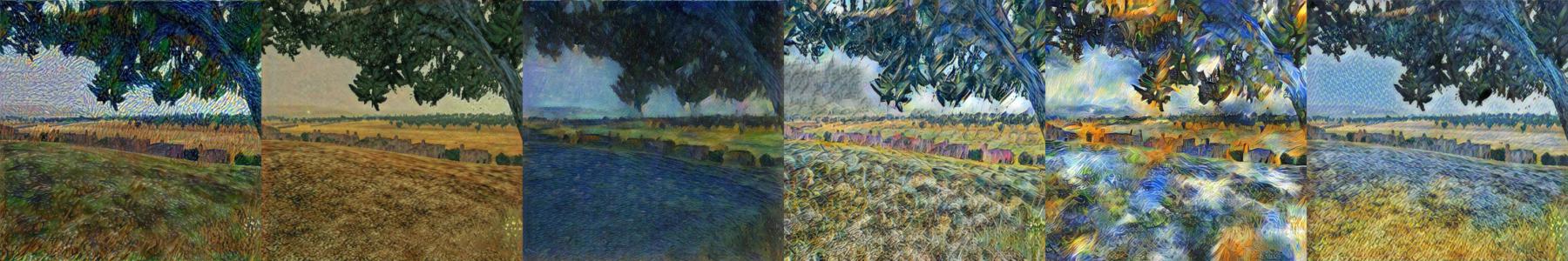}
    \\

    \includegraphics[width=0.11\textwidth,height=0.09\textwidth]{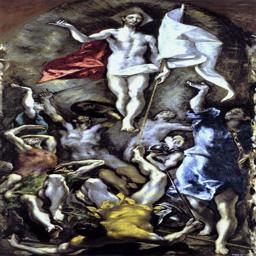} &
    \includegraphics[width=0.11\textwidth,height=0.09\textwidth]{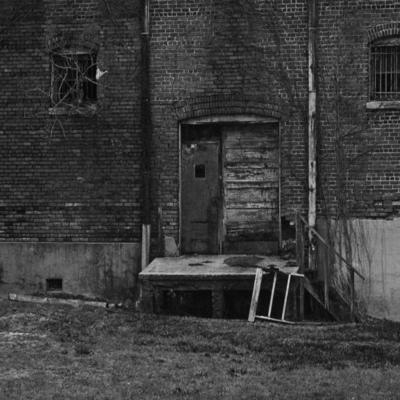} &
    \includegraphics[width=0.11\textwidth,height=0.09\textwidth]{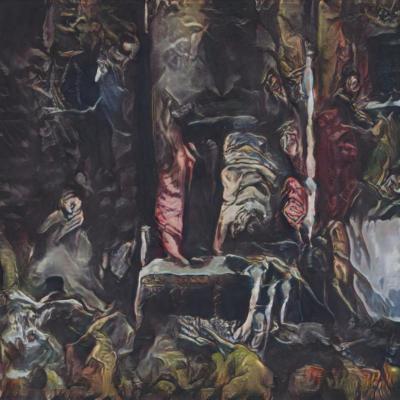} &
    \includegraphics[width=0.66\textwidth,height=0.09\textwidth]{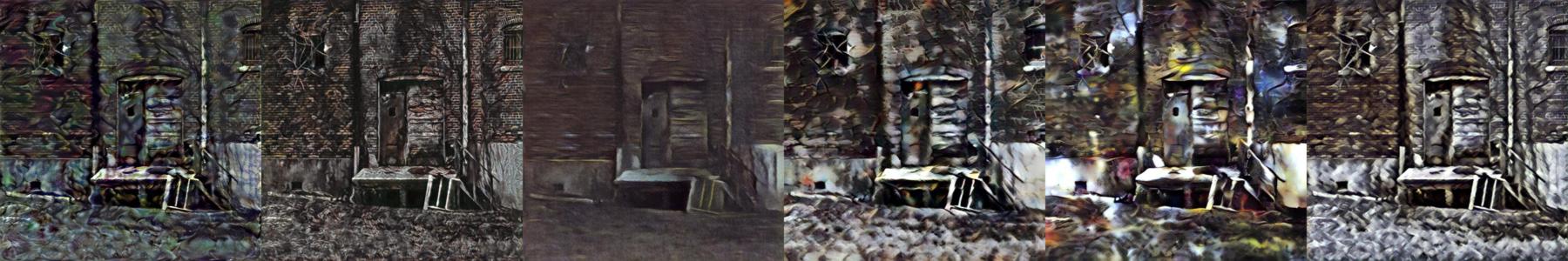}
    \\

    \includegraphics[width=0.11\textwidth,height=0.09\textwidth]{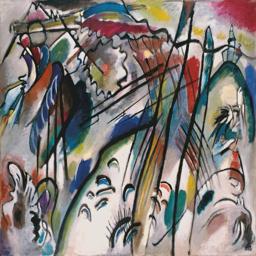} &
    \includegraphics[width=0.11\textwidth,height=0.09\textwidth]{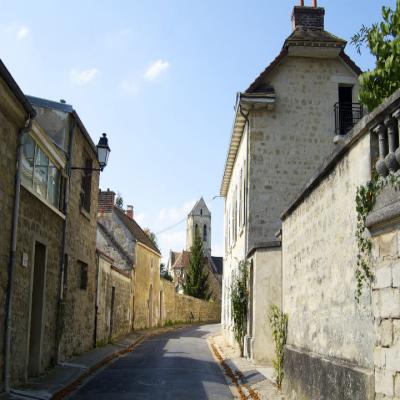} &
    \includegraphics[width=0.11\textwidth,height=0.09\textwidth]{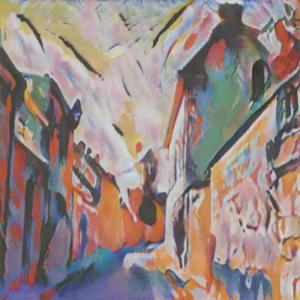} &
    \includegraphics[width=0.66\textwidth,height=0.09\textwidth]{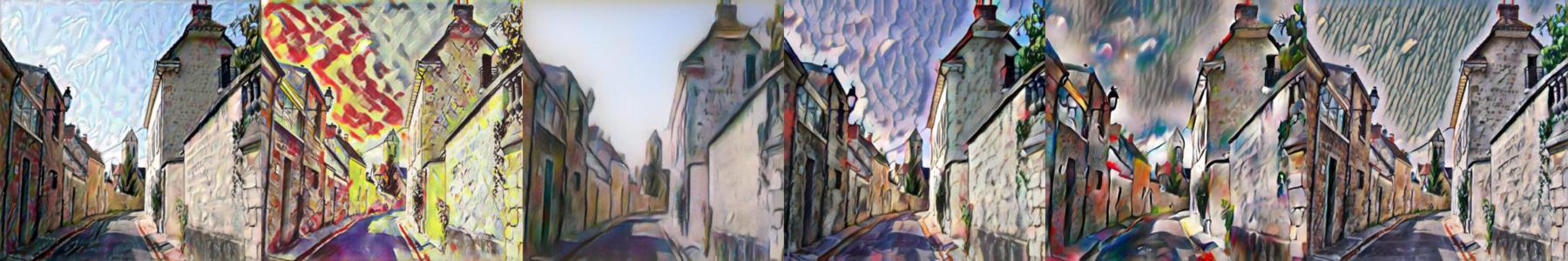}
    \\

    \includegraphics[width=0.11\textwidth,height=0.09\textwidth]{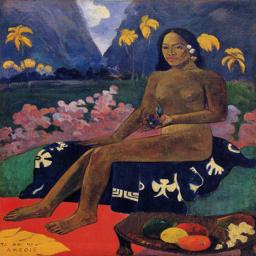} &
    \includegraphics[width=0.11\textwidth,height=0.09\textwidth]{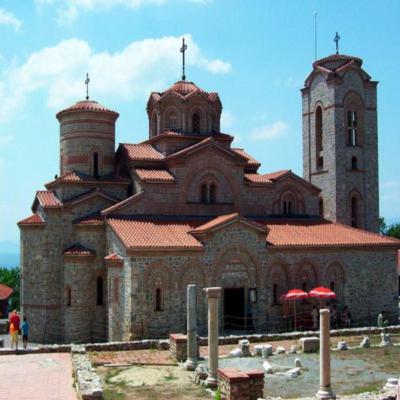} &
    \includegraphics[width=0.11\textwidth,height=0.09\textwidth]{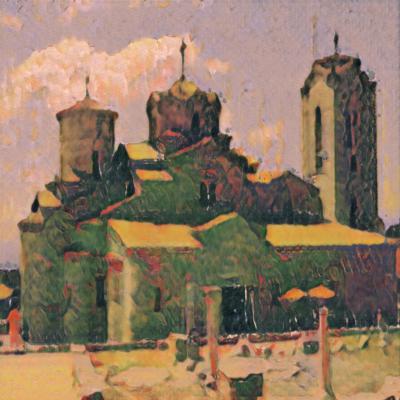} &
    \includegraphics[width=0.66\textwidth,height=0.09\textwidth]{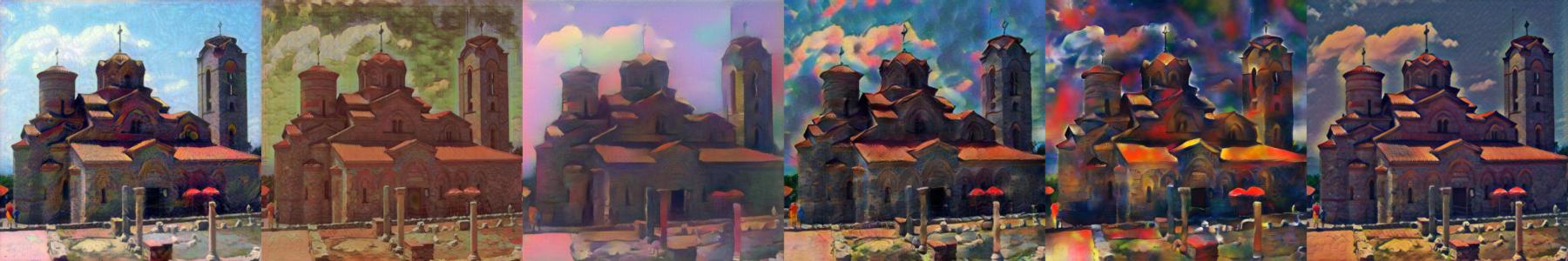}
    \end{tabular}
    \caption{Results from different style transfer methods. We compare methods on different styles and content images.}
    \label{fig:baseline_comparison_full}
\end{figure}

\textbf{Full image stylization:}
In Fig.~\ref{fig:baseline_comparison_full} we demonstrate the effectiveness of our approach for stylizing different contents with various styles.
Chen et al. \cite{pb} work on the overlapping patches extracted from the content image, swapping the features of the original patch with the features of the most similar patch in the style image, and then averages the features in the overlapping regions, thus producing an over-smoothed image without fine details (Fig.~\ref{fig:baseline_comparison_full}(d)).
\cite{adain} produces a lot of repetitive artifacts, especially visible on flat surfaces, cf. Fig.~\ref{fig:baseline_comparison_full}(e, rows 1, 4--6).
Method \cite{universal_style} fails to understand the content of the image and applies different colors in the wrong locations (Fig.~\ref{fig:baseline_comparison_full}(f)). Methods \cite{johnson,cyclegan} often fail to alter content image and their effect may be characterized as shifting the color histogram, e.g., Fig.~\ref{fig:baseline_comparison_full}(g, rows 3, 7; c, rows 1, 3--4). One reason for such failure cases of \cite{cyclegan} is the loss in the RGB pixel space based on the difference between a backward mapping of the stylized output and the content image. Another reason for this is that we utilized the standard Places365 \cite{places365} dataset and did not hand-pick training content images, as is advised for \cite{cyclegan}. Thus, artworks and content images used for training differed significantly in their content, which is the ultimate test for a stylization that truly alters the input and goes beyond a direct mapping between regions of content and style images. The optimization-based method \cite{gatys2016} often works better than other baselines, but produces a lot of prominent artifacts, leading to details of stylizations looking unnatural, cf. Fig.~\ref{fig:baseline_comparison_full}(b, rows 4, 5, 6). This is due to an explicit minimization of the loss directly on the pixel level.
In contrast to this, our model can not only handle styles, which have salient, simple to spot characteristics, but also styles, such as El Greco's Mannerism, with less graspable stylistic characteristics, where other methods fail (Fig.~\ref{fig:baseline_comparison_full}, b--g, $5$th row).
\\[1mm]\textbf{Fine-grained style details:}
In Fig.~\ref{fig:baseline_comparison_patches} we show zoomed in cut-outs from the stylized images.
Interestingly, the stylizations of methods \cite{gatys2016,pb,adain,universal_style,hinton_autoencoder} do not change much across styles (compare Fig.~\ref{fig:baseline_comparison_patches}(d, f--i, rows 1--3)). Zhu et al. \cite{cyclegan} produce more diverse images for different styles, but obviously cannot alter the edges of the content (blades of grass are \emph{clearly visible} on all the cutouts in Fig.~\ref{fig:baseline_comparison_patches}(e)). 
Fig.~\ref{fig:baseline_comparison_patches}(c) shows the stylized cutouts of our approach, which exhibit significant changes from one style to another. Another interesting example is the style of Pollock, Fig.~\ref{fig:baseline_comparison_patches}(row 8), where the style-aware loss allows our model to properly alter content to the point of discarding it -- as would be expected from a Pollock action painting. Our approach is able to generate high-resolution stylizations with a lot of style specific details and retains those content details which are necessary for the style.
\begin{figure}[!t]
    \centering
    \def\arraystretch{0}
    \setlength{\tabcolsep}{-0.0pt}
    \begin{tabular}{c||cc||cc}
    Content & (a) Early period & (b) Stylized & (c) Late period & (d) Stylized\\
   \includegraphics[width=0.196\textwidth]{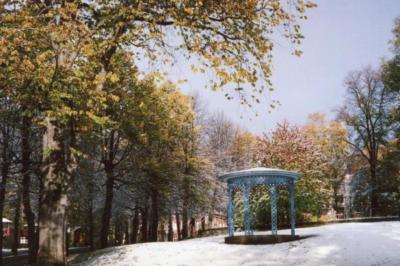} &
   \includegraphics[width=0.196\textwidth]{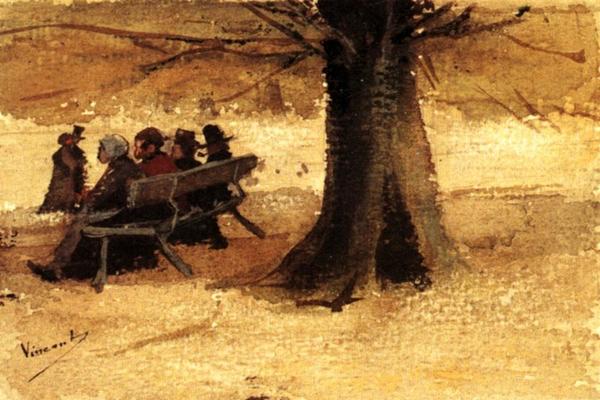} &
   \includegraphics[width=0.196\textwidth]{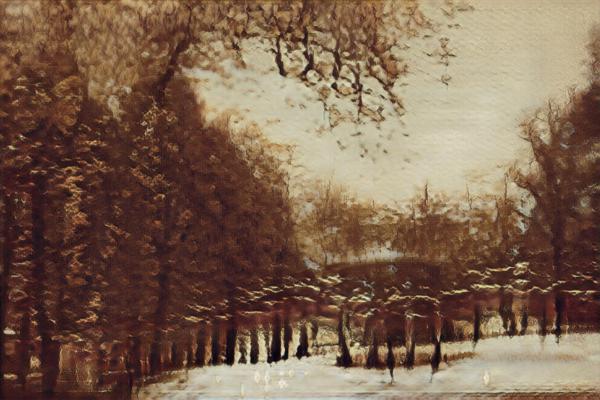} &
   \includegraphics[width=0.196\textwidth]{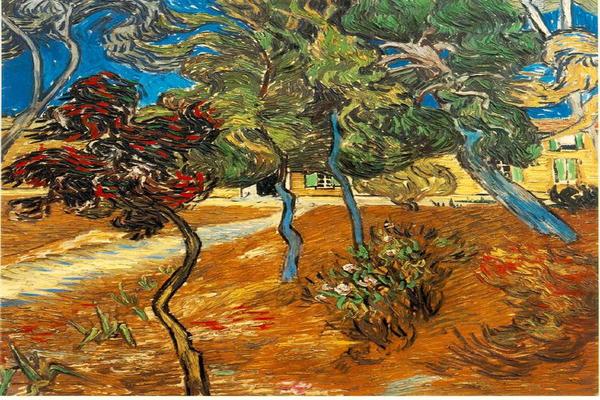}
   &
   \includegraphics[width=0.196\textwidth]{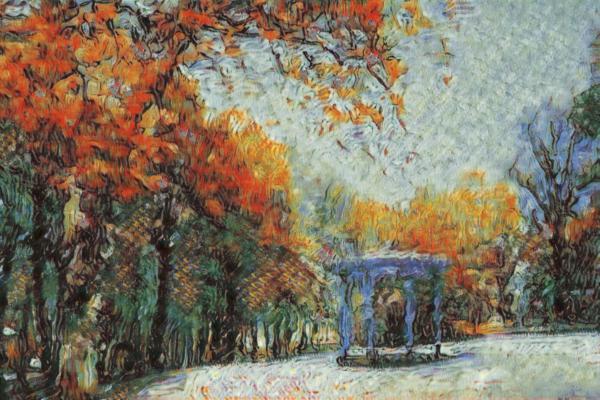}
    \\
    
    \includegraphics[width=0.196\textwidth]{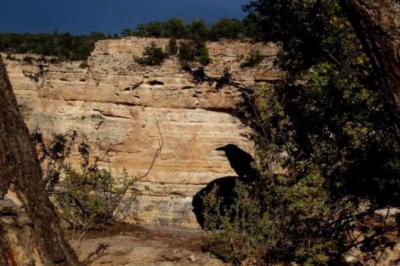} &
   \includegraphics[width=0.196\textwidth]{img/ref_style/early-van-gogh-ref2.jpg} &
   \includegraphics[width=0.196\textwidth]{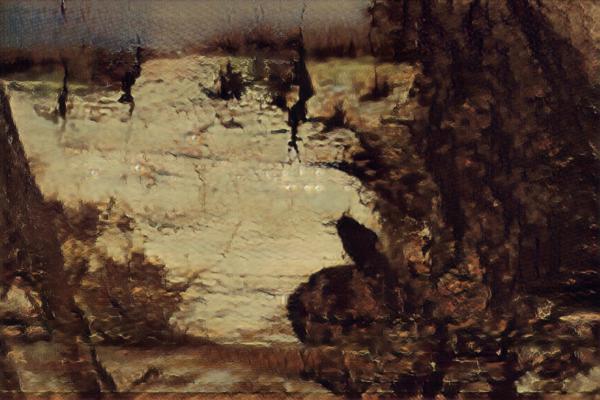} &
   \includegraphics[width=0.196\textwidth]{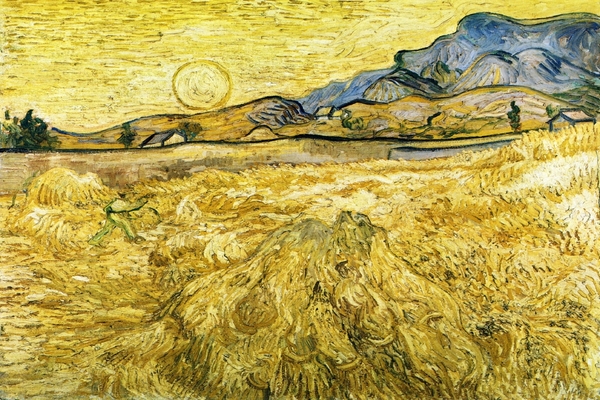}
   &
   \includegraphics[width=0.196\textwidth]{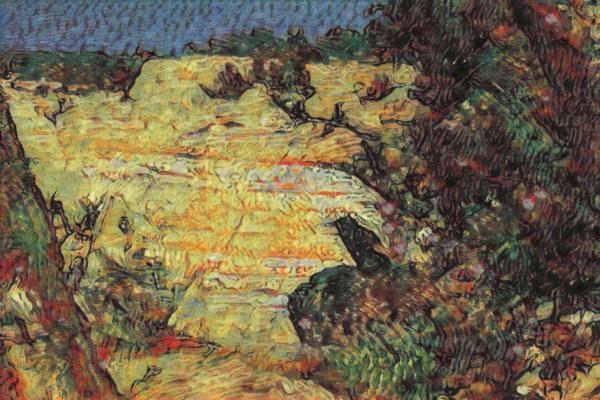}
    \\
    \end{tabular}
    \caption{
    Artwork examples of the early artistic period of van Gogh (a) and his late period (c). Style transfer of the content image (1st column) onto the early period is presented in (b) and the late period in (d).
    }
    \label{fig:early_late_van_gogh}
\end{figure}
\begin{figure}[!t]
    \centering
    \def\arraystretch{0}
    \setlength{\tabcolsep}{-0.0pt}
    
    {
    \scriptsize
    \begin{tabular}{*{3}{C{0.09\textwidth}}*{7}{C{0.09\textwidth}}}
     Style & (a) & (b) & 
     (c) & (d) \cite{gatys2016} & (e) \cite{cyclegan} & 
     (f) \cite{pb} & (g) \cite{adain} & (h) \cite{universal_style} & (i) \cite{johnson}
    \end{tabular}
    }
    
    \begin{tabular}{cc||c||c}
   \includegraphics[width=0.098\textwidth]{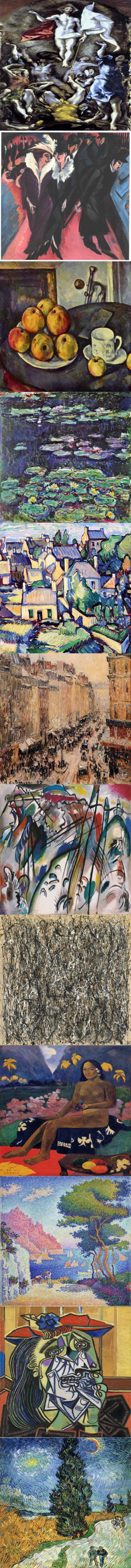} &
   \includegraphics[width=0.098\textwidth]{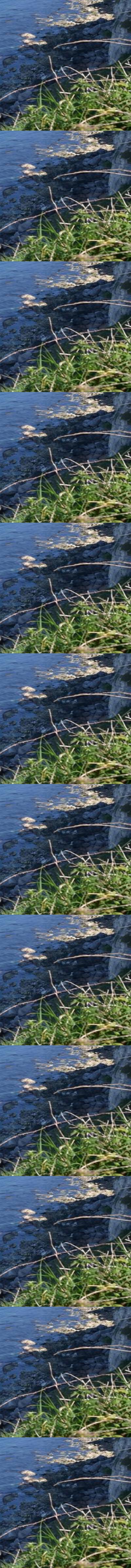} &
   \includegraphics[width=0.098\textwidth]{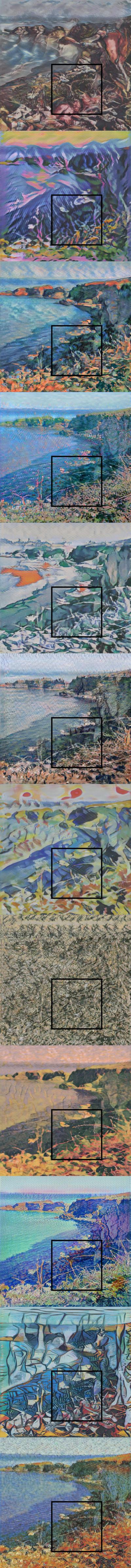} &
   \includegraphics[width=0.686\textwidth]{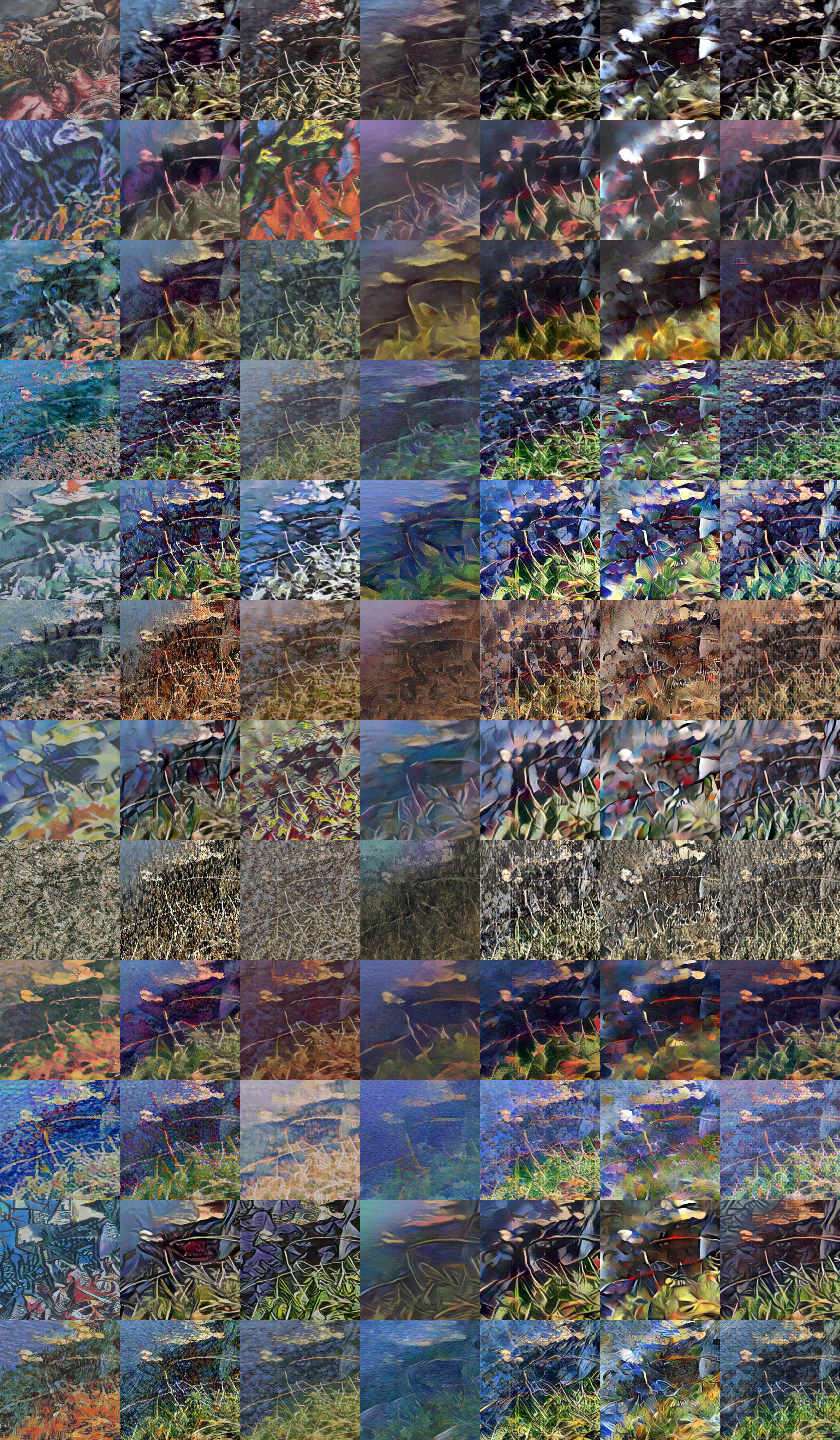}
    \\
    
    \end{tabular}
    \caption{Details from stylized images produced for different styles for a fixed content image (a). (b) is our entire stylized image, (c) the zoomed in cut-out and (d)-(i) the same region for competitors. Note the variation across different styles along the \emph{column} for our method compared to  other approaches. This highlights the ability to adapt content (not just colors or textures) where demanded by a style. Fine grained artistic details with sharp boundaries are produced, while altering the original content edges.}
    \label{fig:baseline_comparison_patches}
\end{figure}
\\[1mm]\textbf{Style transfer for different periods of van Gogh:}
We now investigate our ability to properly model fine differences in style \emph{despite} using a group of style images. Therefore, we take two reference images Fig.~\ref{fig:early_late_van_gogh}(a) and (c) from van Gogh's early and late period, respectively, and acquire related style images for both from Wikiart.
It can be clearly seen that the stylizations produced for either period Fig.~\ref{fig:early_late_van_gogh}(b, d) are fairly different and indeed depict the content in correspondence with the style of early (b) and late (d) periods of van Gogh. This highlights that collections of style images are properly used and do not lead to an averaging effect.
\\[1mm]\textbf{High-resolution image generation:}
Our approach allows us to produce high quality stylized images in high-resolution.
Fig.~\ref{fig:hr} illustrates an example of the generated piece of art in the style of Berthe Morisot with resolution $1280\times1280$. The result exhibits a lot of fine details such as color transitions of the oil paint and brushstrokes of different sizes. More HD images are in the supplementary material.
\begin{figure}[t!]
    \centering
    \includegraphics[width=\textwidth,height=0.8\textwidth]{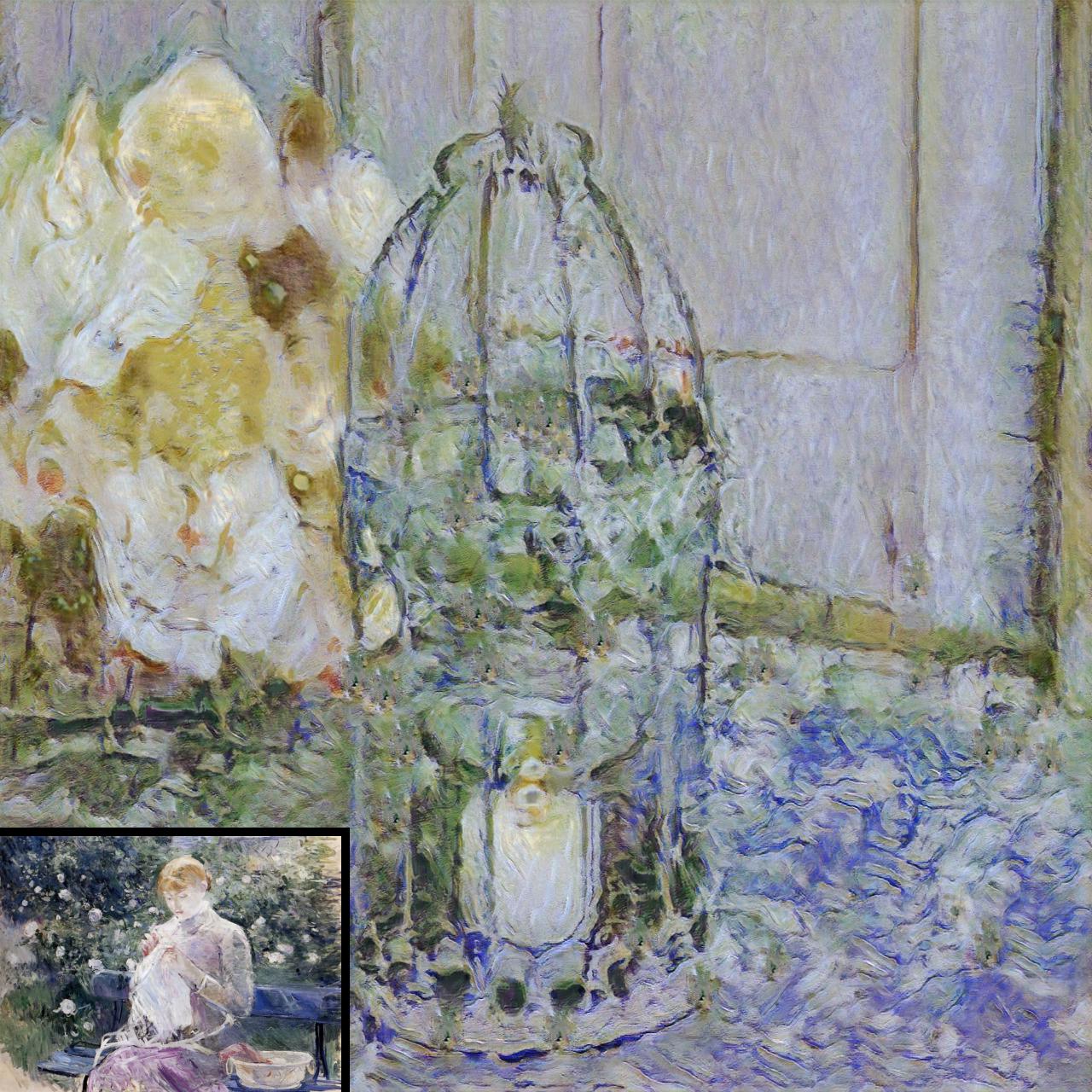}
    \caption{High-resolution image (1280x1280 pix) generated by our approach. A lot of fine details and brushstrokes are visible. A style example is in the bottom left corner.
    }
    \label{fig:hr}
\end{figure}
\\[1mm]\textbf{Real-time HD video stylization:}
We also apply our method to several videos. Our approach can stylize HD videos ($1280 \times 720$) at $9$ FPS. 
Fig.~\ref{fig:video} shows stylized frames from a video. We did not use a temporal regularization to show that our method produces equally good results for consecutive frames with varying appearance w/o extra constraints.
 Stylized videos are in the supplementary material. 

\begin{figure}[!t]
    \centering
    \def\arraystretch{0}
    \setlength{\tabcolsep}{-0.0pt}
    \begin{tabular}{c||c}
    Style examples & Stylized frames
    \\
    \includegraphics[width=0.225\textwidth,height=0.2\textwidth]{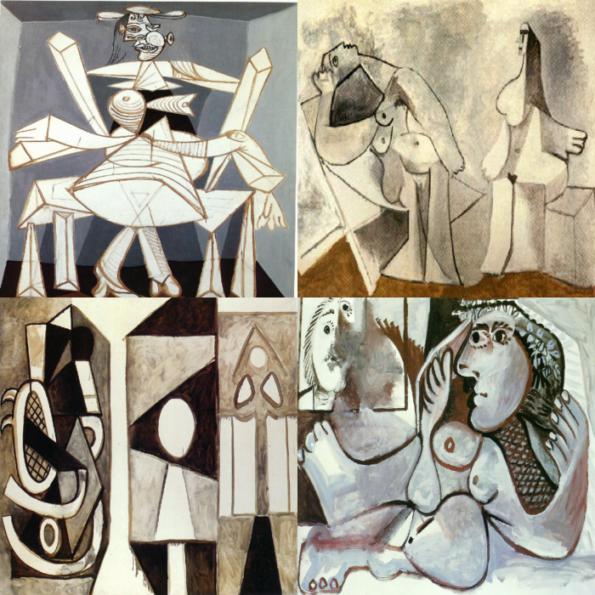} &
    \includegraphics[width=0.7\textwidth,height=0.2\textwidth]{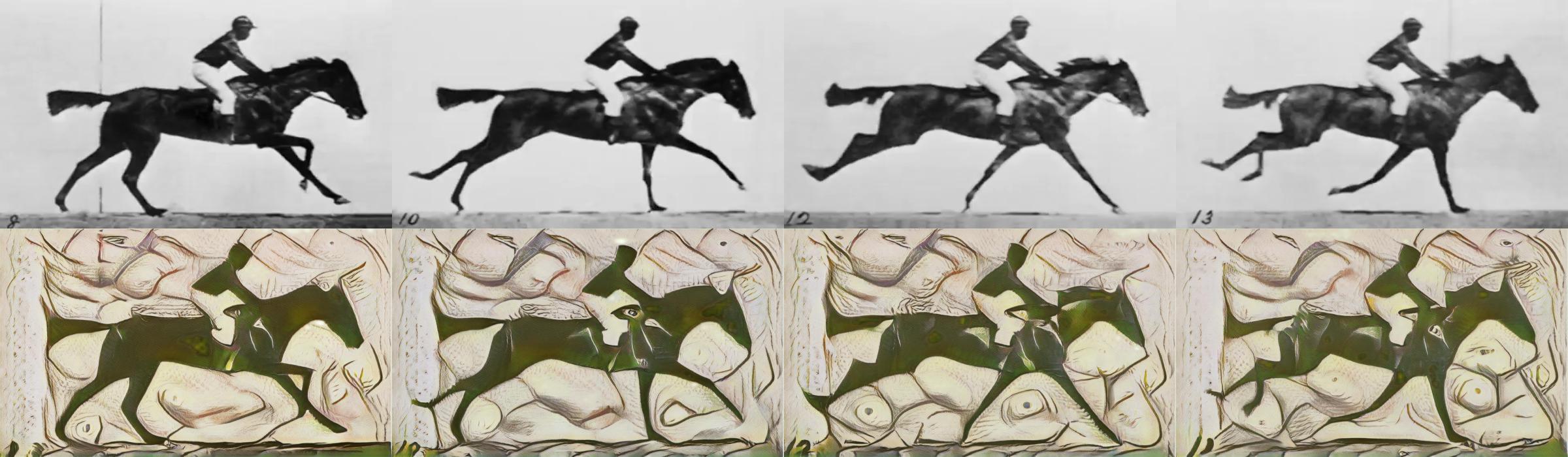}
    \end{tabular}
    \caption{Results of our approach applied to the HD video of Eadweard Muybridge "The horse in motion" (1878). Every frame was \emph{independently}  processed (no smoothing or post-processing) by our model in the style of Picasso.
 Video resolution is $1920 \times 1280$ pix. The full video is available on YouTube: \url{https://youtu.be/TtHJcL8Feu0}.}
    \label{fig:video}
\end{figure}

\subsection{Quantitative Evaluation}
\textbf{Style transfer deception rate:} 
While several metrics \cite{inception_score,mode_score,fid_score} have been proposed to evaluate the quality of image generation, until now no evaluation metric has been proposed for an automatic evaluation of style transfer results. To measure the quality of the stylized images, we introduce the \emph{style transfer deception rate}. 
We use a VGG16 network trained from scratch to classify $624$ artists on Wikiart.
Style transfer deception rate is calculated as the fraction of generated images which were classified by the network as the artworks of an artist for which the stylization was produced.
For fair comparison with other approaches, which used only one style image $y_0$ (hence only one artist), we restricted $Y$ to only contain samples coming from the same artist as the query example $y_0$.
We selected $18$ different artists (i.e. styles).
For every method we generated $5400$ stylizations ($18$ styles, $300$ per style). 
In Tab.~\ref{tab:mean_accuracy} we report mean deception rate for $18$ styles. Our method achieves $0.393$ significantly outperforming the baselines. For comparison, mean accuracy of the network on hold-out real images of aforementioned $18$ artists from Wikiart is $0.616$.
\\[1mm]\textbf{Human art history experts perceptual studies:}  
Three experts (with a PhD in art history with focus on modern and pre-modern paintings) have compared results of our method against recent work. Each expert was shown $1000$ groups of images. Each group consists of stylizations which were generated by different methods based on the same content and style images. Experts were asked to choose one image which best and most realistically reflects the current style. 
The score is computed as the fraction of times a specific method was chosen as the best in the group. We calculate a mean expert score for each method using $18$ different styles and report them in Tab.~\ref{tab:mean_accuracy}. Here, we see that the experts selected our method in around $50\%$ of the cases.
\\[1mm]\textbf{Speed and memory:}
Tab.~\ref{tab:speed_memory} shows the time and memory required for stylization of a single image of size $768\times768$ px for different methods.
One can see that our approach and that of \cite{johnson} and \cite{cyclegan} have comparable speed and only very modest demands on GPU memory, compared to modern graphics cards.

\begin{table}
\begin{minipage}{.49\linewidth}
    \scriptsize
    \caption{Mean deception rate and mean expert score for different methods. The higher the better.}
    \centering
    \vfill
    \begin{tabular}{lcc}
    \hline
    \noalign{\smallskip}
    Method & Deception rate & Expert score\\
    \noalign{\smallskip}
    \hline
    \noalign{\smallskip}
    Content images & 0.002 & -\\
    AdaIn~\cite{adain} & 0.074 & 0.060\\
    PatchBased~\cite{pb} & 0.040 & 0.132\\
    Johnson et al.~\cite{johnson} & 0.051 & 0.048\\ 
    WCT~\cite{universal_style} & 0.035 & 0.044 \\
    CycleGan~\cite{cyclegan} & 0.139 & 0.044\\
    Gatys et al.~\cite{gatys2016} &  0.147 & 0.178 \\
    \textbf{Ours} & \textbf{0.393} & \textbf{0.495} \\
    \hline
    \end{tabular}
    \label{tab:mean_accuracy}
\end{minipage}
\hfill
\begin{minipage}{.45\linewidth}
    \scriptsize
    \caption{{Average inference time and GPU memory consumption, measured on a Titan X Pascal, for different methods with batch size~$1$ and input image of $768\times 768$ pix.}}
    \centering
    \begin{tabular}{l r r}
    \hline\noalign{\smallskip}
    Method & Time & GPU memory\\
    \noalign{\smallskip}
    \hline
    \noalign{\smallskip}
    
    Gatys et al.~\cite{gatys2016} & 200 sec & 3887 MiB\\
    CycleGan~\cite{cyclegan} & 0.07 sec & 1391 MiB\\
    AdaIn~\cite{adain} & 0.16 sec & $8872$ MiB\\
    PatchBased~\cite{pb} & 8.70 sec & 4159 MiB\\
    WCT~\cite{universal_style} & $5.22$ sec & $10720$ MiB \\
    Johnson et al.~\cite{johnson} & $\textbf{0.06}$ \textbf{sec} & \textbf{$\textbf{671}$ MiB}\\ 
    \textbf{Ours} & $0.07$ sec & $1043$ MiB \\
    \hline
      
    \end{tabular}
    \label{tab:speed_memory}
\end{minipage}
\end{table}

\subsection{Ablation Studies}
\label{sec:ablation}

\begin{figure}[t!]
    \centering
    \def\arraystretch{0}
    \setlength{\tabcolsep}{0.0pt}
    \begin{tabular}{ccccccc}
    Style & (a) Content & (b) Ours & (c) & (d) & (e) & (f) \\
    \includegraphics[width=0.1\textwidth,height=0.1125\textwidth]{img/ref_style/Gauguin-the-seed-of-the-areoi-1892.jpg} &
    \includegraphics[width=0.15\textwidth]{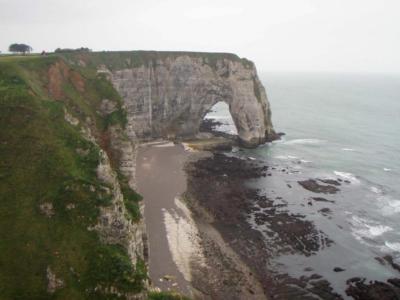} &
    \includegraphics[width=0.15\textwidth]{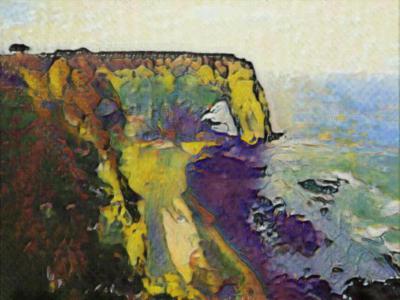} & 
    \includegraphics[width=0.15\textwidth]{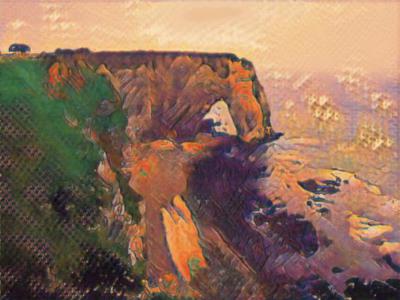} &
    \includegraphics[width=0.15\textwidth]{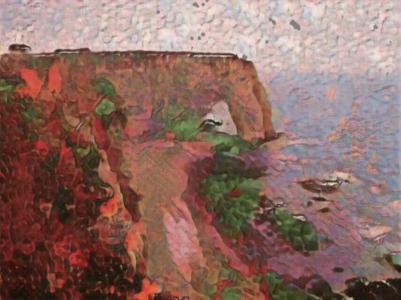} &
    \includegraphics[width=0.15\textwidth]{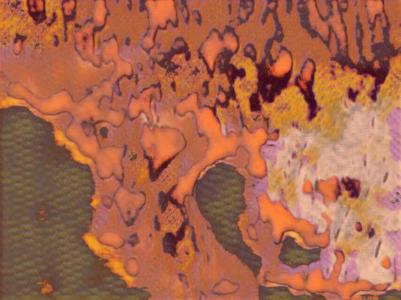} &
    \includegraphics[width=0.15\textwidth]{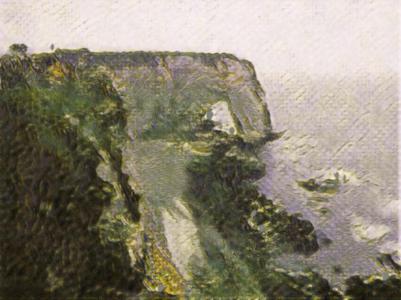} \\
    
    \end{tabular}
    \caption{Different variations of our method for Gauguin stylization. See Sect.~\ref{sec:ablation} for details.
    (a) Content image; (b) full model ($\mL_c$, $\mL_{rgb}$ and $\mL_D$); 
    (c) $\mL_{rgb}$ and $\mL_D$; 
    (d) without transformer block;
    (e) only $\mL_D$;
    (f) trained with all of Gauguin's artworks as style images. Please zoom in to compare.}

    \label{fig:ablation_no_feat}
\end{figure}

\textbf{Effect of different losses:}
We study the effect of different components of our model in Fig.~\ref{fig:ablation_no_feat}.
Removing the style-aware content loss significantly degrades the results, (c). We observe that without the style-aware loss training becomes instable and often stalls. If we remove the transformed image loss, which we introduced for a proper initialization of our model that is trained from scratch, we notice mode collapse after $5000$ iterations. 
Training directly with pixel-wise L$2$ distance causes a lot of artifacts (grey blobs and flaky structure), (d). Training only with a discriminator neither exhibits the variability in the painting nor in the content, (e). Therefore we conclude that both the style-aware content loss and the transformed image loss are critical for our approach. 
\begin{figure}[!t]
    \centering
    \def\arraystretch{0.0}
    \setlength{\tabcolsep}{-0.0pt}
    \begin{tabular}{cccc}
    Style & Content & (a) & (b)\\
    \includegraphics[width=0.23\textwidth,height=0.18\textwidth]{img/ref_style/vincent-van-gogh_road-with-cypresses-1890.jpg} &
    \includegraphics[width=0.23\textwidth,height=0.18\textwidth]{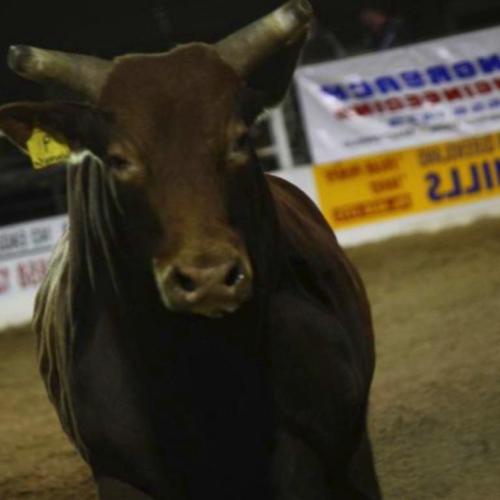} &
    \includegraphics[width=0.23\textwidth,height=0.18\textwidth]{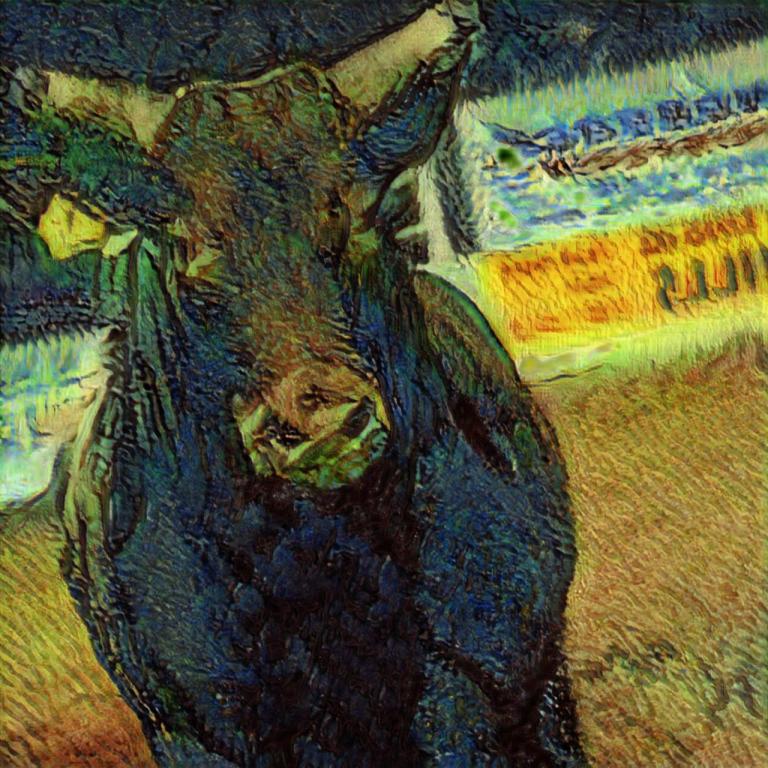} &
    \includegraphics[width=0.23\textwidth,height=0.18\textwidth]{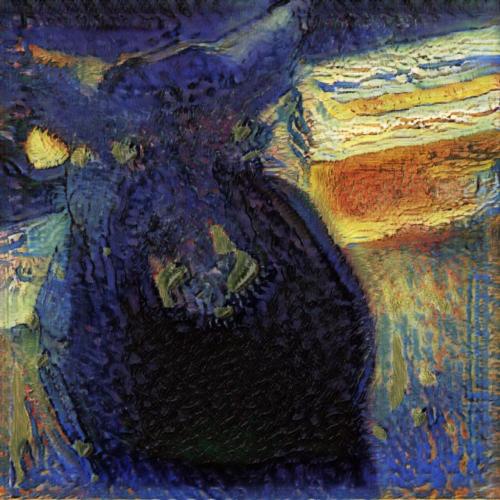}
    \end{tabular}
    \caption{Encoder ablation studies: (a) stylization using our model; (b) stylization using pre-trained VGG16 encoder instead of $E$.}
    \label{fig:vgg_encoder}
\end{figure}
\\[1mm]\textbf{Single vs collection of style images:}
Here, we investigate the importance of the style image grouping. First, we trained a model with only one style image of Gauguin, which led to mode collapse.
Second, we trained with all of Gauguin's artworks as style images (without utilizing style grouping procedure). It produced unsatisfactory results, cf. Fig.~10%
(f), because style images comprised several distinct styles. Therefore we conclude that to learn a good style transfer model it is important to group style images according to their stylistic similarity.
\\[1mm]\textbf{Encoder ablation:} To investigate the effect of our encoder $E$, we substitute it with VGG16 \cite{vgg} encoder (up to \texttt{conv5\_3}) pre-trained on ImageNet. The VGG encoder retains features that separate object classes (since it was trained discriminatively), as opposed to our encoder which is trained to retain style-specific content details. Hence, our encoder is not biased towards class-discriminative features, but is style specific and trained from scratch. Fig.~\ref{fig:vgg_encoder}(a, b) show that our approach produces better results than with pre-trained VGG16 encoder.
\let\svthefootnote\thefootnote
\let\thefootnote\relax\footnotetext{\label{trick:solution}\textit{Solution to Fig.~\ref{fig:trick}: patches $3$ and $5$ were generated by our approach, others by artists.}}

\section{Conclusion}
This paper has addressed major conceptual issues in state-of-the-art approaches for style transfer. We overcome the limitation of only a single style image or the need for style and content training images to show similar content. Moreover, we exceed a mere pixel-wise comparison of stylistic images or models that are pre-trained on millions of ImageNet bounding boxes. The proposed style-aware content loss enables a real-time, high-resolution encoder-decoder based stylization of images and videos and significantly improves stylization by capturing how style affects content.
\footnote{
This work has been supported in part by a DFG grant, the Heidelberg Academy of Science, and an Nvidia hardware donation. 
}
\addtocounter{footnote}{-1}\let\thefootnote\svthefootnote 

\clearpage

\bibliographystyle{splncs04}
\bibliography{egbib}

\newpage
\appendix

\section{Implementation Details}

\subsection{Network Architecture}
We incarnate our approach using an encoder-decoder 
architecture with a discriminator.
Below we follow the naming convention similar to the one used in \cite{cyclegan}.

Let:
\begin{itemize}[*,topsep=0pt]
    \item \texttt{cFs1-k} denote $F \times F$ Convolution-InstanceNorm-ReLU layer with $k$ filters and stride $1$;
    \item \texttt{cFs1-k-sigmoid} denote $F \times F$ Convolution-InstanceNorm-Sigmoid layer with $k$ filters and stride $1$;
    \item \texttt{cFs1-k-noact} denote $F \times F$ Convolution-InstanceNorm layer with $k$ filters and stride $1$;
    \item \texttt{dF-k} denote a $F \times F$ Convolution-InstanceNorm-ReLU layer with $k$ filters and stride $2$;
    \item \texttt{dF-k-LReLU} denote a $F \times F$ Convolution-InstanceNorm-LeakyReLU layer with $k$ filters and stride $2$;
    \item \texttt{Rk} denote a residual block that contains two $3 \times 3$ convolutional layers followed by InstanceNorm \cite{instanceNorm};
    \item  \texttt{uk} denote an upscaling block that contains nearest-neighbor upscaling by a factor of $2$ followed by Convolution-InstanceNorm-ReLU layer with $k$ filters and stride $1$.
\end{itemize}
All convolutional layers use reflection padding.

\subsubsection{Encoder-decoder architecture}
Encoder has InstanceNorm layer at the beginning and $5$ convolutional layers:
\texttt{InstanceNorm, c3s1-32, d3-32, d3-64, d3-128, d3-256.}
Decoder contains 9 residual blocks, $4$ upscaling blocks and $1$ convolution $7 \times 7$ with sigmoid activation: 
\texttt{R256$\times 9$, u256, u128, u64, u32, c7s1-3-sigmoid.}

\subsubsection{Transformer block $\mathbf{T}$ }
Transformer block $\mathbf{T}$ contains a convolutional layer followed by a weight normalization layer with fixed norm $||W||=1$. Convolutional layer: $3$ kernels of size $10 \times 10$, stride $1$, uniformly initialized.

\subsubsection{Discriminator architecture}
Discriminator is implemented as a fully convolutional network and tries to classify if input image patches are real or fake.
It contains $7$ convolutional layers. We use Leaky ReLU activations with slope $0.2$.
Architecture is defined as:
\texttt{d5-128-LReLU, d5-128-LReLU, d5-256-LReLU, d5-512-LReLU, d5-512-LReLU, d5-1024-LReLU, d5-1024-LReLU,\\ c3s1-1-noact}.

We use $4$ auxiliary classifiers (implemented as a convolutional layer with $1$ filter each) to capture image details on different scales \cite{iizuka2017globally,xue2017segan} and to alleviate artifacts: auxiliary classifiers were added after 1st, 2nd, 4th and 6th convolutional layers of the discriminator. We simply sum up all the losses from different scales.

\subsection{
Style Image Grouping Details
}
As described in Sec.~\ref{sec:grouping}, the number of style images is not fixed and depends on the neighborhood size of the query image $y_0$ and how frequent the style is in the Wikiart \cite{karayev2013recognizing} dataset. For different style examples in our experiments it varied from $55$ to $1391$ related images. We observe: the more style examples the better, as long as they are from the same style. Ablation studies (Sec.~\ref{sec:ablation}) show that using too few style examples (e.g. one) leads to mode collapse; at the same time using a lot of images, which are less related to each other (e.g. all images of an artist), produces unsatisfactory stylizations.

\begin{figure}[t!]
    \centering
    \def\arraystretch{0.0}
    \setlength{\tabcolsep}{-0.0pt}
    \begin{tabular}{c}
    \includegraphics[width=1.0\textwidth]{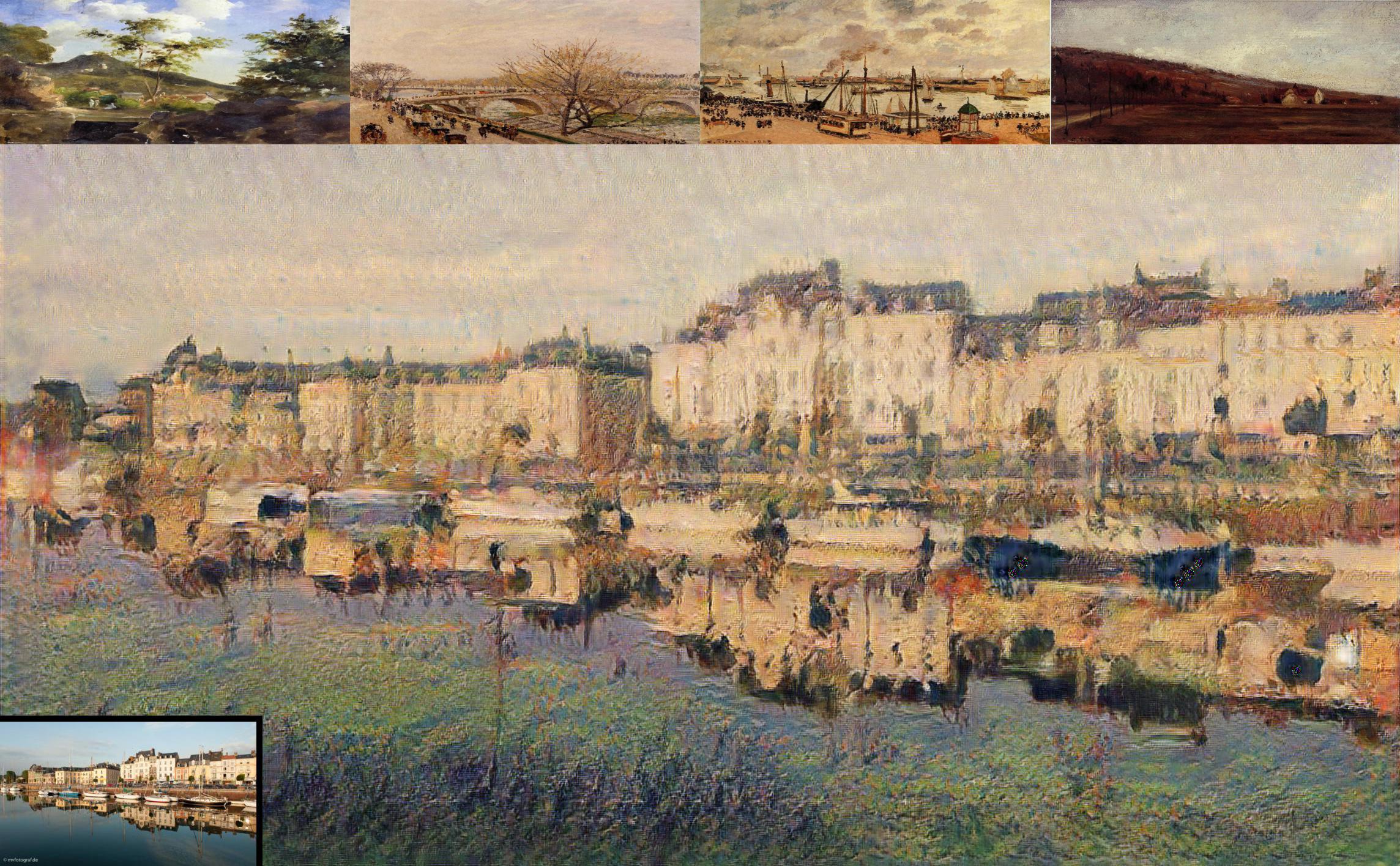} 
    \end{tabular}
    \caption{High-resolution ($2288\times1280$ pix) Pissarro stylization by our approach. The top row depicts $4$ randomly chosen from set $Y$ instances of style  obtained by the style image grouping.}
    \label{fig:hr5}
\end{figure}
\begin{figure}[t!]
    \centering
    \def\arraystretch{0.0}
    \setlength{\tabcolsep}{-0.0pt}
    \begin{tabular}{c}
 
    \includegraphics[width=1.0\textwidth]{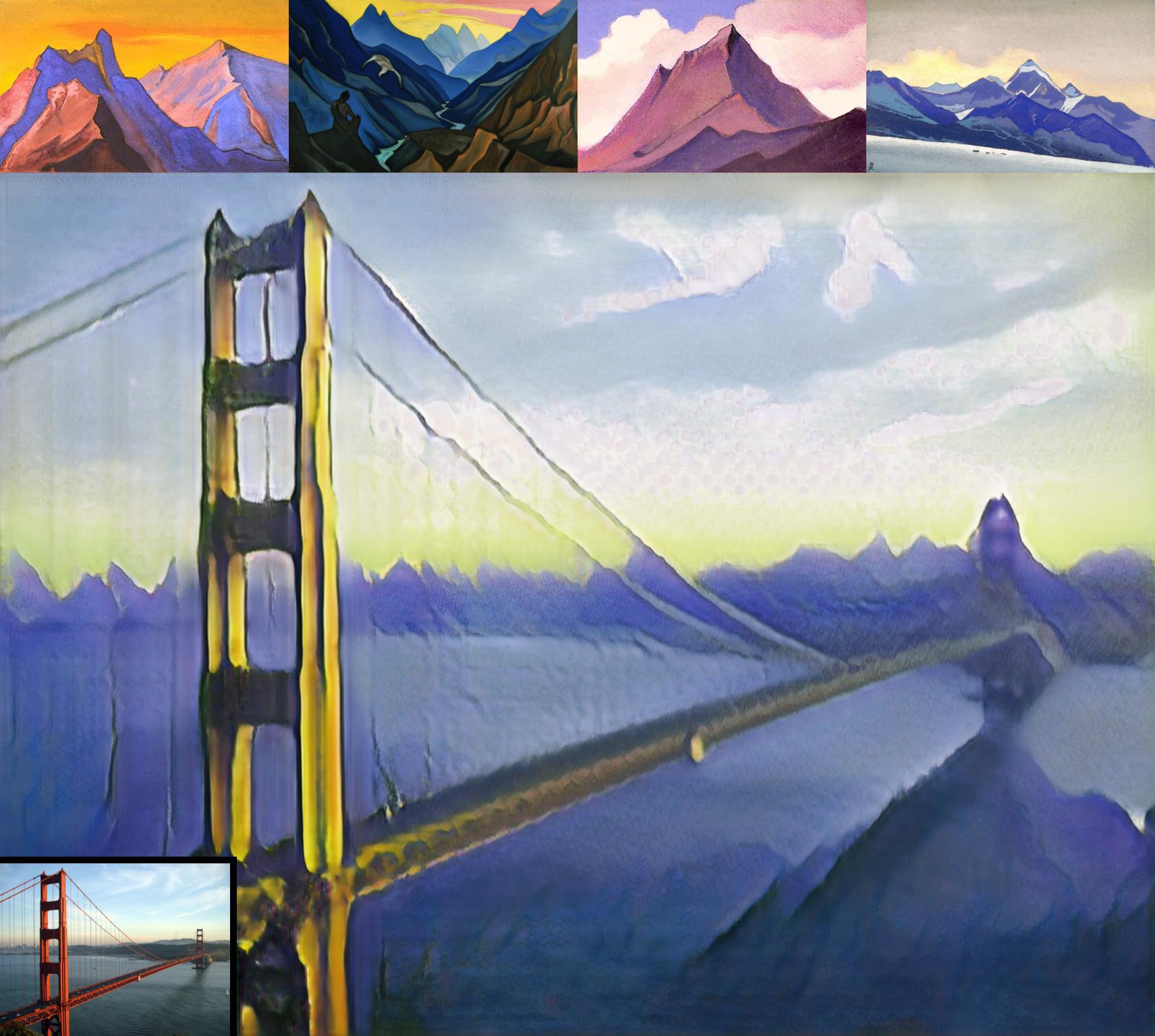} 
    \end{tabular}
    \caption{High-resolution ($1712\times1280$ pix) Roerich stylization  by our approach. The top row depicts $4$ randomly chosen from set $Y$ instances of style  obtained by the style image grouping.}
    \label{fig:hr9}
\end{figure}

\subsection{Training Details}
All networks were trained from scratch on randomly cropped image patches of size $768 \times 768$ pix. $\lambda$ in Eq.~\ref{eq:total_loss} was set to $0.001$.
We trained for $300000$ iterations with batch size $1$ and learning rate $2 \times 10^{-4}$ using Adam \cite{adam} optimizer. We reduced the learning rate by a factor of $10$ after $200000$ iterations. 

 In each iteration we alternatively update encoder-decoder and discriminator.
To balance out the discriminator and encoder-decoder training \cite{gans_overwiev} we update solely the discriminator, if it has accuracy $< 0.8$, and update only encoder-decoder otherwise. The discriminator accuracy is calculated using exponential moving average during training.

\section{Extra Qualitative Results}

\subsubsection{Altering the style of an existing artwork}
Our method is able to change the style of an existing artwork, rendering it in another style. It means, that our algorithm can also handle content images which are artistic, which, to our knowledge, was never shown in previous work before. We refer the reader to the results on the project page\footref{footnote:project_url}.

\subsubsection{Real-time HD video stylization}
We stylized Eadweard Muybridge's "The horse in motion" video, 1878, in the style of Picasso. The stylized video resolution is $1920 \times 1280$ pix. 
Additionally, we apply our method to the video "Aix-en-Provence, France"  
by Jan van Bekkum in resolution $1920 \times 1280$ pix. We stitch together the original video and $3$ stylizations in the style of C\'ezanne, Kandinsky and Picasso produced by our approach. The videos are available on the project page\footref{footnote:project_url}.

\subsubsection{More examples of high-resolution style transfer}
In Fig.~\ref{fig:hr5}, \ref{fig:hr9} we show extra high-resolution images stylized by our approach.
Additional comparison of our method against baselines and more high-resolution visual results are available on the project page\footref{footnote:project_url}.

\section{Extra ablation studies}
\begin{figure}[!t]
    \centering
    \def\arraystretch{0.0}
    \setlength{\tabcolsep}{-0.0pt}
    \begin{tabular}{cccc}
    (a) Style & (b) Content & (c) Ours & (d)\\
     \includegraphics[width=0.23\textwidth,height=0.18\textwidth]{img/ref_style/vincent-van-gogh_road-with-cypresses-1890.jpg} &
    \includegraphics[width=0.23\textwidth,height=0.18\textwidth]{img/Rebuttal/bull_500_500.jpg} &
    \includegraphics[width=0.23\textwidth,height=0.18\textwidth]{img/Rebuttal/compare_to_VGG_encoder/trained_encoder/bull_stylized_vincent-van-gogh_768_768.jpg} &
    \includegraphics[width=0.23\textwidth,height=0.18\textwidth]{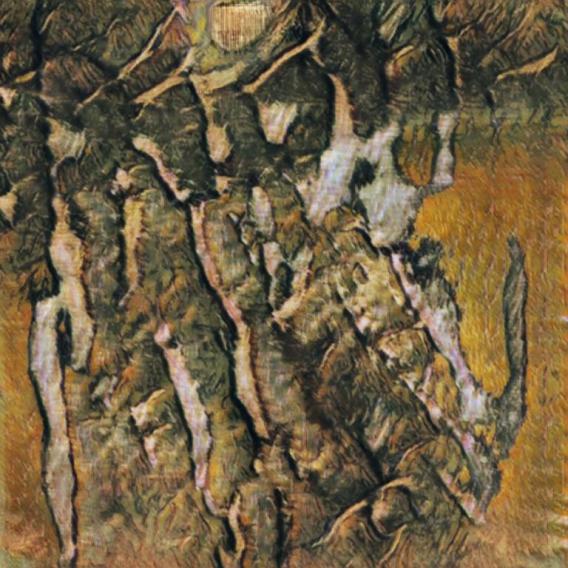}
    \end{tabular}
    \caption{Investigation of the importance of an independent transformer block. (a) style image (van Gogh); (b) content image; (c) our stylization; (d) stylization with the loss $\mL_{conv1}$ instead of $\mL_{\mathbf{T}}$.}
    \label{fig:vgg_encoder_appendix}
\end{figure}

\subsubsection{Transformer block ablation}
To demonstrate the need in an independent transformer block,
we replaced the transformed image loss $\mL_{\mathbf{T}}$ with a loss $\mL_{conv1}$ using \texttt{conv1} features  of the encoder,
\begin{equation}
\mL_{conv1}=(E, G) = \mathop{\mathbb{E}}_{x \sim p_{X}(x)} \left[ \frac{1}{d_{conv1}} || \phi_{conv1}(x) - \phi_{conv1}(G(E(x)) ||_2^2 \right],
\end{equation}
where $\phi_{conv1}(x)$ are the activations of the \texttt{conv1} layer of the encoder $E$ for input image $x$ and $d_{conv1}$ is the dimensianality of $\phi_{conv1}(x)$.
This 
produced worse results (cf. Fig.~\ref{fig:vgg_encoder_appendix}(d)), since the loss $\mL_{conv1}$ is then directly tied to the convolutional layer used for our style-aware content representation in the encoder.
In contrast, the proposed transformer block can be learned independently from the early layers of the encoder.

\end{document}